\pdfoutput=1

\documentclass[11pt]{article}
\usepackage[preprint]{acl}
\usepackage{amsmath}
\usepackage{times}
\usepackage{latexsym}
\usepackage{placeins}
\usepackage{amsfonts}
\usepackage{colortbl}
\usepackage{multirow}
\usepackage{booktabs}
\usepackage{arydshln}
\usepackage{caption}
\usepackage{hhline}
\usepackage{amssymb}
\usepackage{makecell}
\usepackage{xcolor}
\usepackage{xspace}
\usepackage{hyperref}
\usepackage{subcaption}
\usepackage{listings}
\usepackage{comment}
\usepackage{float}
\usepackage[T1]{fontenc}
\usepackage{tikz}
\usepackage[most]{tcolorbox}

\usepackage[utf8]{inputenc}

\usepackage{natbib}

\usepackage{microtype}
\usepackage{inconsolata}

\usepackage{graphicx}

\definecolor{aliceLPBlue}{HTML}{2F80ED}
\definecolor{aliceKEGreen}{HTML}{2EAD67}
\definecolor{aliceSKYellow}{HTML}{D9A600}
\newcommand{\tasklp}{\textcolor{aliceLPBlue}{LP}\xspace}
\newcommand{\taskke}{\textcolor{aliceKEGreen}{KE}\xspace}
\newcommand{\tasksk}{\textcolor{aliceSKYellow}{SK}\xspace}
\newcommand{\alicelp}{\textsc{Alice}-\textcolor{aliceLPBlue}{LP}\xspace}
\newcommand{\aliceke}{\textsc{Alice}-\textcolor{aliceKEGreen}{KE}\xspace}
\newcommand{\alicesk}{\textsc{Alice}-\textcolor{aliceSKYellow}{SK}\xspace}
\title{\textsc{Alice}: A Large-Scale German Benchmark for Rubric-Based Multidimensional Automatic Short Answer Scoring}

\author{
  \textbf{Zhifan Sun\textsuperscript{1}},
  \textbf{Sebastian Gombert\textsuperscript{1}},
  \textbf{Jannik Lossjew\textsuperscript{2}},
  \textbf{Tobias Wyrwich\textsuperscript{2}},
  \\
  \textbf{Berrit Katharina Czinczel\textsuperscript{2}},
  \textbf{David Bednorz\textsuperscript{2}},
  \textbf{Marcus Kubsch\textsuperscript{3}},
  \\
  \textbf{Knut Neumann\textsuperscript{2}},
  \textbf{Hendrik Drachsler\textsuperscript{1,4}}
  \\
  \\
  \textsuperscript{1}DIPF | Leibniz Institute for Research and Information in Education
  \\
  \textsuperscript{2}IPN | Leibniz Institute for Science and Mathematics Education
  \\
  \textsuperscript{3}Ume\r{a} University
  \\
  \textsuperscript{4}Computer Science Department \& Studiumdigitale, Goethe University Frankfurt
  \\
  \small\texttt{\{z.sun,s.gombert,h.drachsler\}@dipf.de}
  \\
  \small\texttt{\{wyrwich,lossjew,czinczel,bednorz,neumann\}@leibniz-ipn.de}
  \\
  \small\texttt{marcus.kubsch@umu.se}
}

\begin{document}
\maketitle
\begin{abstract}

Automatic short answer scoring (ASAS) is central to NLP for education.
However, openly available benchmarks remain scarce, and existing datasets largely address how well students answer
a question directly rather than how well they master underlying \textbf{concepts} (knowledge elements) 
such as \textit{thermal energy}
or \textbf{epistemic activities} (skills) such as reasoning or claim formulation.
To address this gap, we introduce \textsc{Alice}, a German rubric-based ASAS benchmark with annotations for three subtasks:
(i) learning performance (\alicelp),
(ii) knowledge elements (\aliceke), and (iii) skills (\alicesk).
We compare sequence classification, zero-shot LLM prompting, and rubric-retrieval, which serves as our primary benchmark formulation.
Our experiments show that \textsc{Alice} is challenging across all three subtasks, especially on the novel \aliceke and \alicesk subtasks.
While sequence classification remains competitive on \alicelp, rubric-retrieval performs more consistently on
\aliceke and \alicesk.
Zero-shot LLMs generally trail supervised approaches, particularly on the more fine-grained subtasks.%
\footnote{The dataset, code, and reference implementations are available at \url{https://github.com/Szhifan/emnlp2026alice}.}

\end{abstract}

\section{Introduction}

\begin{figure}[!t]
    \centering
    \includegraphics[width=0.8\linewidth]{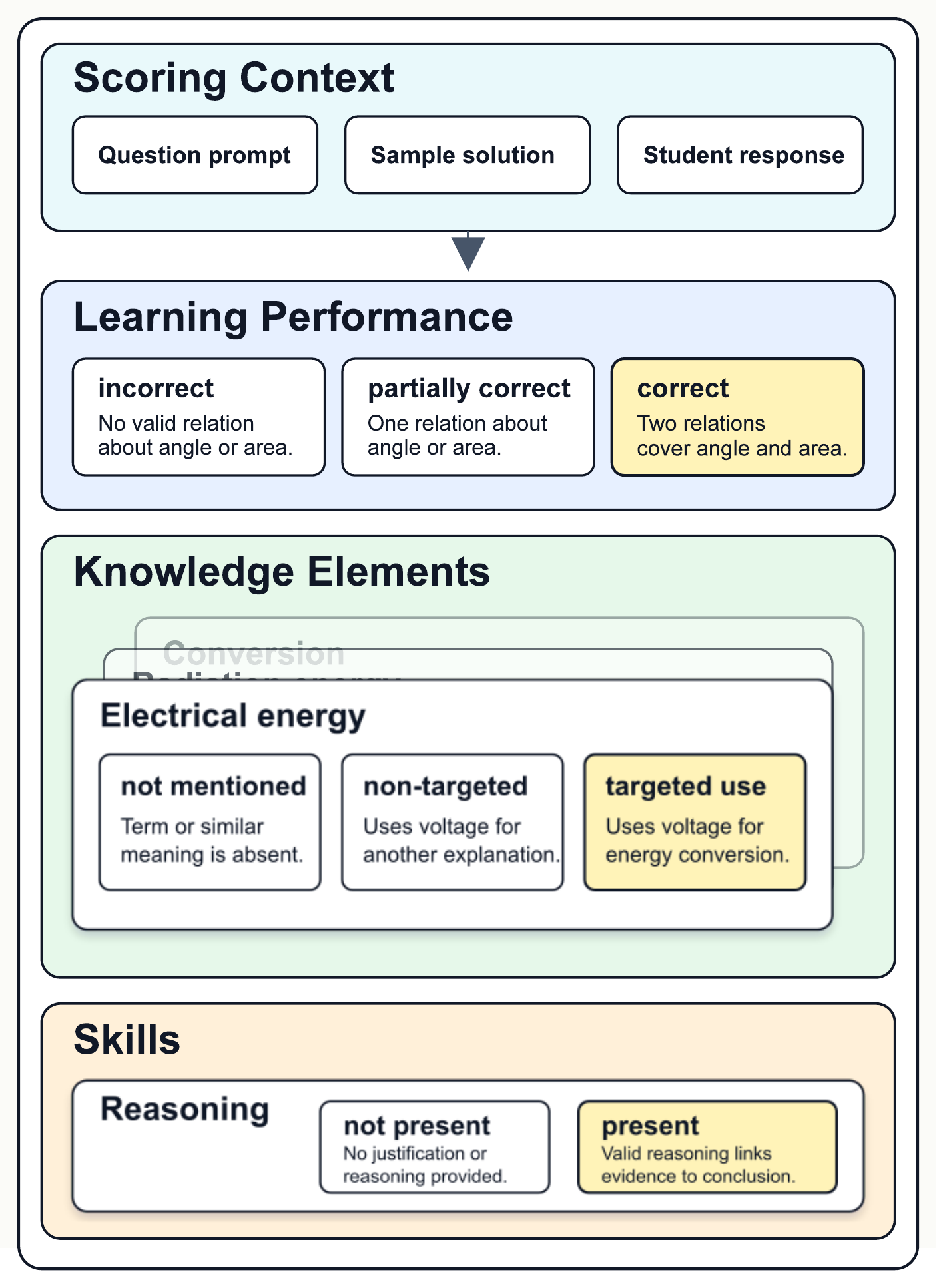}
    \caption{An instance from \textsc{Alice}. Each instance includes a question prompt, a sample solution, and per-level rubrics for overall learning performance, knowledge elements, and skills. The complete example is shown in \autoref{fig:dataset_instance} in the appendix.}
    \label{fig:dataset}
\end{figure}

Automatic short answer scoring (ASAS) is a fundamental task in NLP for education.
Extensive research has focused on both algorithmic approaches~
\cite{bexte-etal-2022-similarity,li-etal-2023-answer-state, zehner-etal-2025-cascades, wang-etal-2019-inject-rubric, Bexte2024strengthandweakness}
and the development of benchmark datasets \citep{dzikovska-etal-2013-semeval, safe}.
In real-world educational settings, especially in scientific education, short-answer questions are designed to probe students' understanding of concepts taught during instruction.
By constructing answers of their own, students can actively reflect on prior learning
and foster critical thinking~\cite{bai_stede_2023_student_eval_survey}.
As manually grading students' answers is tedious and expensive, developing benchmarks and models that automate this process can significantly alleviate the workload of
teachers and ensure timely and individualised feedback. Despite substantial progress, existing ASAS benchmarks and models remain largely misaligned with how human teachers
evaluate student responses.

First, most benchmarks, and consequently the models trained and evaluated on them, adopt a solution-based formulation,
in which scoring is cast as a comparison with the solution, and they often assume a uniform set of levels across
questions~\cite{domain-pretrain-asas, safe, asas-transformer, ormerod2022shortanswerscoringensemblespretrained, kumar2019get}.
In practice, however, questions may have different performance levels depending on pedagogical needs,
and teachers rely on question-specific rubrics~\citep{brookhart2018appropriatecriteria, panadero_jonsson_2013_scoring_rubrics_formative_review, krebs2022rubrics} for scoring.

Second, in terms of scoring objectives, knowledge elements (concepts) and skills (epistemic activities) are critical in scientific education, where a central goal of pedagogical activities is for students
to master scientific concepts and apply epistemic activities to solve
problems~\citep{assessment-student-concept, coding_energy_knowledge, reynders_lantz_ruder_stanford_cole_2020_rubrics_process_skills_stem}.
Yet, existing ASAS benchmarks only evaluate how students address questions directly.

To address the gaps, we expand \alicelp~1.1, introduced by \citet{gombert2026bea_shared_task}, into \textsc{Alice}, a multidimensional German rubric-based ASAS benchmark.
Compared with the \alicelp~1.1 release, the version used here contains additional student responses and includes annotations for three subtasks: \alicelp, which measures how well students answer the question;
\aliceke, which measures students' mastery of the target concept; and \alicesk, which evaluates students' epistemic activities.
This setting exposes the limitations of previously predominant sequence-classification approaches, which assume a uniform set of possible levels across all questions.

For modelling, we use rubric-retrieval as a primary benchmark formulation,
in which student responses are scored by aligning
them with question-specific rubric items rather than predicting fixed labels.
This formulation provides a simple and practical benchmarking setup for variable,
question-specific rubrics and is aligned with the benchmark design.

Crucially, every instance in \textsc{Alice} is accompanied by the question prompt, sample solution,
and per-level rubrics for overall learning performance and for each knowledge element and skill
(\autoref{fig:dataset}), closely aligned with how human teachers evaluate responses.
This design allows us to test the extent to which rubric-based ASAS models rely on these contextual
components to score responses effectively (see~\autoref{appendix:instance} for a full instance).

The contributions of our work can be summarised as follows:
\begin{itemize}
  \item We evaluate rubric-based ASAS models by expanding \alicelp~1.1 with additional student responses and \taskke/\tasksk annotations under a newly defined split.
  \item We provide a benchmark setup for rubric-based ASAS by approaching scoring as retrieval over question-specific rubric candidates, and compare this setup against sequence classification and zero-shot prompting baselines.
  \item We present an empirical analysis across masked language models (MLMs) and large language models (LLMs) to study the effect of rubric information and optional context on performance within this expanded setting. We find that MLMs are more prone to performance degradation when given additional question and solution context, whereas LLM encoders benefit from this context more often, suggesting that they can use richer rubric context more reliably.
\end{itemize}
\section{Related Work} \label{sec2:related_work}

\subsection{Short Answer Scoring: Benchmarks}
\autoref{tab:asas_comparison} provides an overview of existing ASAS benchmarks.
Publicly available ASAS benchmarks, especially non-English ones, remain scarce. Existing benchmarks predominantly focus on how well student responses address the question in general, rather than on the mastery of underlying concepts or skills. \textsc{SciEntsBank}~\cite{dzikovska-etal-2013-semeval} and \textsc{ASAP-SAS}\footnote{\url{https://www.kaggle.com/competitions/asap-sas}} are the most widely used datasets in prior work. Multilingual benchmarks include the German \textsc{SAF}~\cite{safe}, the Portuguese \textsc{PT\_ASAG\_2018}~\cite{galhardi2018portuguese}, and the Japanese \textsc{RIKEN-SAS} dataset~\citep{mizumoto-etal-2019-analytic, funayama2025crossprompt}. \citet{sonkar2024automatedlonganswergrading} extend this setting to paragraph-length answers.
Due to privacy constraints, much work in this field evaluates scoring models on proprietary datasets~\cite{chang2024finnish, domain-pretrain-asas, zehner-etal-2025-cascades}.

\begin{table*}[t]
    \centering
    \small
    \resizebox{1\textwidth}{!}{
    \begin{tabular}{l r r r r r l}
        \hline
        \textbf{Benchmark} & \textbf{Answers} & \textbf{Questions} & \textbf{Context information} & \textbf{Language coverage} & \textbf{Public?} & \\
        \hline
        \textsc{ASAP-SAS} & 22K & 10 & question prompt, per-level rubrics & English & $\checkmark$ & \\
        \textsc{SciEntsBank}~\cite{dzikovska-etal-2013-semeval} & 10K & 197 & solution, question prompt & English & $\checkmark$ & \\
        \textsc{Beetle}~\cite{dzikovska-etal-2013-semeval} & 3K & 56 & solution, question prompt & English & $\checkmark$ & \\
        \textsc{PT\_ASAG\_2018}~\cite{galhardi2018portuguese} & 13K & 8 & solution, question prompt & Portuguese & $\checkmark$ & \\
        \textsc{RIKEN-SAS}~\citep{mizumoto-etal-2019-analytic, funayama2025crossprompt} & 31K & 34 & \parbox[t]{3.5cm}{reading passage, question prompt, analytic rubrics, key phrases} & Japanese & $\checkmark$ & \\
        \citet{domain-pretrain-asas} & 76K & 28 & solution, question prompt & English & $\times$ & \\
        \textsc{SAF}~\cite{safe} & 4.5K & 30 & solution, question & German, English & $\checkmark$ & \\
        \textsc{IStudio}~\cite{li-etal-2023-answer-state} & 6.5K & 6 & \parbox[t]{3.5cm}{question context, question prompt, per-level reference answer} & English & $\checkmark$ & \\
        \citet{chang2024finnish} & 76K & 10 & question prompt & Finnish & $\times$ & \\
        \textsc{RiceChem}~\citep{sonkar2024automatedlonganswergrading} & 1.2K & 4 & additive rubrics & English & $\checkmark$ & \\
        \textsc{SAS-Bench}~\cite{lai2025sasbenchfinegrainedbenchmarkevaluating} & 4.1K & 1K & additive rubrics & Chinese & $\checkmark$ & \\
        \textsc{Alice}-\textsc{LP}~1.1~\cite{gombert2026bea_shared_task} & 13K & 117 & \parbox[t]{3.5cm}{question prompt, solution, per-level LP rubrics} & German & $\checkmark$ & \\
        \textsc{Alice} (this work) & 16K & 112 & \parbox[t]{3.5cm}{question prompt, solution, per-level rubrics for LP, KE, and SK} & German & $\checkmark$ & \\
        \hline
    \end{tabular}
}
    \caption{Comparison of the \textsc{Alice} version used in this work with existing ASAS benchmarks. Note that we frame all information other than the student's answer as \textit{context information}.}
    \label{tab:asas_comparison}
\end{table*}

\paragraph{Rubrics} We define rubrics in the context of ASAS as natural-language descriptions of student answers at each performance level.
Most ASAS benchmarks listed above are solution-based, scoring the student's answer by comparison with the reference answer. While straightforward to implement, solution-based scoring only measures similarity to the highest-level answer and gives little signal for incorrect or partially correct answers. Such responses often reveal valuable insights into students' learning progress~\cite{sadler1989formative, Fisher1986TwentyQA}.

Among publicly available rubric-based ASAS benchmarks, \textsc{ASAP-SAS}
and \textsc{Alice} assume an \textbf{exclusive} relation among rubrics; i.e.,
one answer can satisfy only one rubric. By contrast, \textsc{RiceChem}~\cite{sonkar2024automatedlonganswergrading} and \textsc{SAS-Bench}~\cite{lai2025sasbenchfinegrainedbenchmarkevaluating} assume an \textbf{additive}
relation among rubrics, where the final score is obtained by summing the scores of matched rubrics.

Although \textsc{RIKEN-SAS} provides level-specific rubrics, prior modelling work on the dataset uses rubric-provided key phrases as compact positive references for each analytic criterion and regresses a score from the key phrases and student answer~\citep{mizumoto-etal-2019-analytic, funayama2025crossprompt}.
This design relies on selected reference expressions rather than directly comparing the answer against the full set of level-specific rubric descriptions.
\subsection{Short Answer Scoring: Modelling}
Early ASAS approaches used engineered features such as
answer length and lexical overlap~\cite{dzikovska-etal-2013-semeval, asag_eras_trends}. With the advent of \textit{BERT}~\cite{devlin2019bertpretrainingdeepbidirectional}, the field has
largely shifted toward pretrained language models. A typical approach is to jointly encode the answer and sample solution, casting
scoring as similarity or entailment prediction between the answer and the solution~\cite{domain-pretrain-asas, asas-transformer}. Parallel to this,
a small number of works have incorporated rubric information to support scoring, such as \citet{wang-etal-2019-inject-rubric} and \citet{sonkar2024automatedlonganswergrading}.

Recently, a growing number of works have explored LLM prompting
for ASAS~\cite{chang2024finnish, ferreira2025automatic, lai2025sasbenchfinegrainedbenchmarkevaluating}. Although this approach avoids task-specific fine-tuning,
high performance is usually achieved through large proprietary models, which induce high inference costs and latency. In addition, as a classification task, ASAS does not require free-form generation.

Moreover, ASAS is fundamentally an NLU task: the goal is to classify or rank rubric levels given the answers and context information.
Framing it as generative decoding is computationally inefficient, since each prediction requires token-by-token
generation.
In contrast, encoder-based models produce alignment scores in a single forward pass, resulting in substantially lower inference cost.
\section{The \textsc{Alice} Dataset} \label{sec:3}
\subsection{Collection and Definition}

\textsc{Alice} expands \alicelp~1.1, the German rubric-based learning-performance dataset introduced by \citet{gombert2026bea_shared_task}.
The \alicelp~1.1 release focuses on learning-performance scoring, whereas the version used in this work contains additional student responses and augments the prior learning-performance annotations
with fine-grained scoring of knowledge elements and skills.
Each question is associated with three types of rubrics:
\begin{itemize}
\item \textbf{Learning Performance (\tasklp)}: the degree to which a student's response directly addresses the question prompt. The three performance levels are \textit{incorrect}, \textit{partially correct}, and \textit{correct}.
\item \textbf{Knowledge Elements (\taskke)}: the extent to which a student's response correctly employs the targeted domain concepts specified for the question. Possible levels are \textit{No use}, \textit{Use without content}, \textit{Non-targeted use}, and \textit{Targeted use}. The \textit{Use without content} level is unavailable for some questions.
\item \textbf{Skills (\tasksk)}: the presence of specified cognitive or reasoning behaviours in a student's response. The usual levels are \textit{Not present} and \textit{Present}. For some questions, we further define \textit{Partially present}.
\end{itemize}
These dimensions were chosen to reflect complementary aspects of assessment commonly used in scientific education: task completion,
conceptual understanding, and epistemic activity. Note that a question can contain multiple target \taskke{}s and \tasksk{}s, whereas mathematics questions do not contain
\taskke{}s or \tasksk{}s. The same \taskke or \tasksk items can be shared across multiple questions, but their rubrics may vary for different question contexts.

Each instance in \textsc{Alice} is represented as a tuple $\langle q, a, s, R_{lp}^{(q)}, \{R_t^{(q)}\}_{t \in T_q^{\mathrm{ke}}}, \{R_t^{(q)}\}_{t \in T_q^{\mathrm{sk}}} \rangle$, where $q$ denotes the question prompt,
$a$ the student response, $s$ a sample solution, $R_{lp}^{(q)}$ is the learning-performance rubric set for question $q$, and the final two components are the collections of item-specific rubric sets for its knowledge elements and skills.

The structure of the rubric sets differs across subtasks. For \textbf{\alicelp}, the rubric set
\[
R_{lp}^{(q)} = \{ r^{(q)}_0, r^{(q)}_1, r^{(q)}_2 \}
\]
corresponds to the ordered performance levels defined for question $q$, where each rubric item describes the qualitative criteria for one score level.

For \textbf{\aliceke} and \textbf{\alicesk}, each question $q$ contains a set of \taskke/\tasksk items:
\[
T_q = \{ t_1, \dots, t_m \},
\]
where each \taskke/\tasksk item $t_j$ is associated with its own rubric set $R^{(q)}_{t_j}$,
which defines the scoring criteria for that specific concept or skill. The number of levels in \taskke and \tasksk rubrics can vary across questions. The \taskke and \tasksk items, as well as the rubrics,
are designed by pedagogical experts in the respective subjects.

\subsection{Annotation}\label{sec3:dataset-creation}

The data collection and \tasklp annotation protocol follows \citet{gombert2026bea_shared_task}.
For the extended \taskke and \tasksk annotations, the scoring dimensions themselves were designed by 
educators specialising in each subject, and the rest of the annotation workflow was similar to that of \alicelp~1.1.
We used the same general pilot-and-annotation workflow, with expert annotators scoring responses in four subject-specific phases on the \textsc{INCEPTION}~\cite{klie-etal-2018-inception} platform.
During the pilot rounds, annotator feedback was used to refine question-specific guidelines and rubric wording.
Agreement scores are reported in \autoref{tab:qwk_subject_agreement} as quadratic weighted kappa ($\kappa$), computed separately for each subject and scoring dimension rather than aggregated into a single overall value.

\begin{table}[t]
\centering
\resizebox{0.4\linewidth}{!}{%
\setlength{\tabcolsep}{6pt}
\renewcommand{\arraystretch}{1.12}
\begin{tabular}{lcc}
\toprule
\textbf{Subject} & \textbf{\taskke} & \textbf{\tasksk} \\
\midrule
Biology     & 0.84 & 0.78 \\
Chemistry   & 0.86 & 0.65 \\
Physics     & 0.72 & 0.69 \\
\bottomrule
\end{tabular}
}
\caption{Quadratic weighted kappa ($\kappa$; QWK) agreement for \taskke and \tasksk annotations by subject.}
\label{tab:qwk_subject_agreement}
\end{table}

\subsection{Semantic Overlap between Rubric Levels} \label{subsection:rubentail}

Rubric levels in \textsc{Alice} typically form a progression from absence of evidence to increasingly complete
demonstrations of understanding. For instance, a higher level requires that the student do \textit{X and Y}, while a lower one requires that the student do \textit{X or Y}.
Higher-level descriptions might subsume requirements expressed at lower levels rather than describing
contradictory requirements. As a result, an answer satisfying a higher-level rubric may
also fulfil parts of a lower-level description, even though only one level constitutes the final score.

To examine how many rubric pairs display this structural property,
we apply natural language inference (NLI) to pairs of rubrics from adjacent levels, where we treat the rubric of a higher level as the premise
and the rubric of a lower level as the hypothesis. The experimental setup is described in \autoref{appendix:rubentail}.
\begin{figure}[t]
    \centering
    \includegraphics[width=0.8\columnwidth]{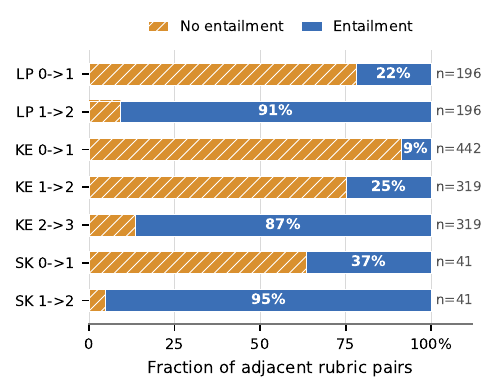}
    \caption{NLI label distribution between adjacent higher- and lower-level rubric pairs.}
    \label{fig:rubentail}

\end{figure}
The results are shown in \autoref{fig:rubentail}. We can see that across all three subtasks, adjacent levels do not usually contradict each other, but are often semantically
overlapping.
\subsection{Dataset Statistics}\label{sec3:dataset-statistics}
\begin{figure*}[t]
    \centering
    \includegraphics[width=0.7\textwidth]{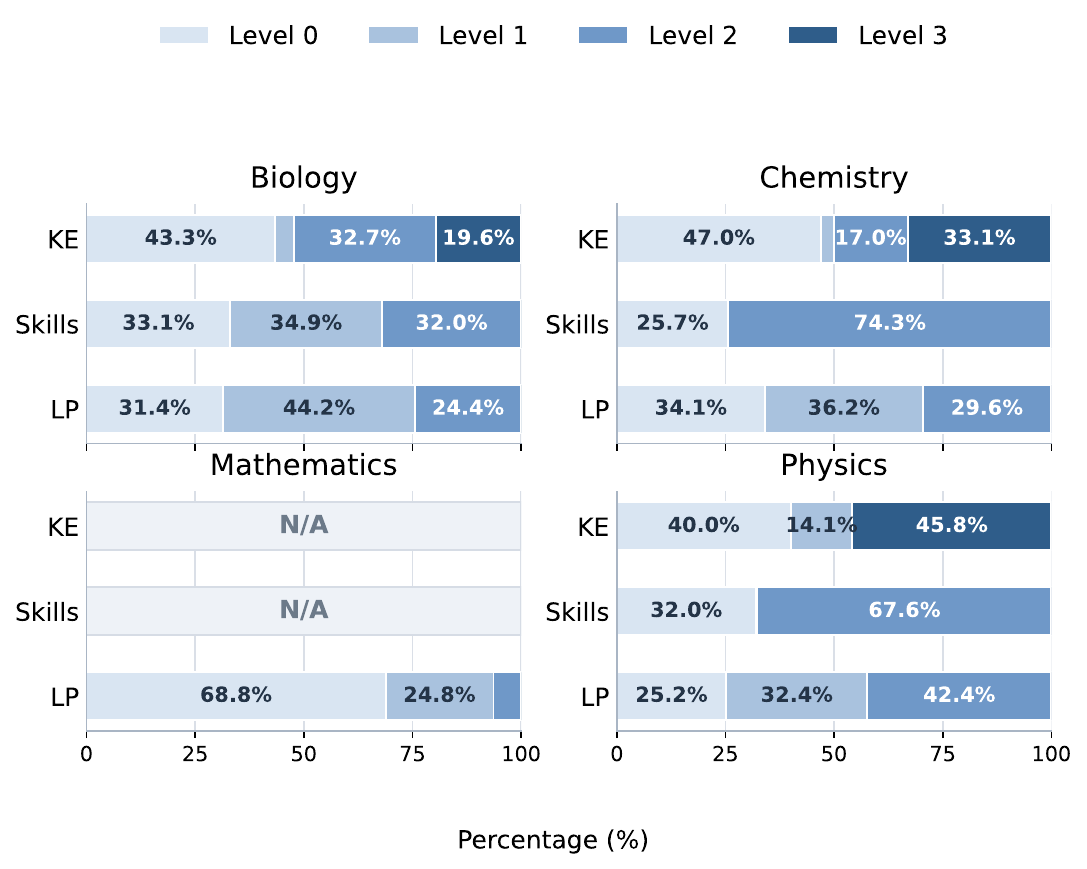}
    \caption{Distribution of levels across subtasks and subjects. Mathematics has no \taskke or \tasksk annotations.}
    \label{fig:label-dis}
\end{figure*}

After annotation, we filter out questions without sample solutions (e.g., open-ended questions) to normalise the dataset structure.
The result is a corpus of 16,572 student answers. To ensure robust generalisation, we use a newly defined split with Test-UA (unseen answers) and
Test-UQ (unseen questions). We construct this split in two steps: first, 22 of the 112 questions are sampled uniformly at random and held out entirely as Test-UQ, so that none of their answers appear in training; the remaining 90 questions form the question pool from which Train and Test-UA are drawn at the answer level. This split differs from the \alicelp~1.1 split because the dataset version used here contains additional student responses and \taskke/\tasksk annotations. Consequently, the results reported in this paper are not directly comparable to published baseline results associated with that release. Detailed split statistics and label distributions by subject are reported in \autoref{tab:stats} and \autoref{fig:label-dis}.
\begin{table*}[t]
\centering
\small
\setlength{\tabcolsep}{5pt}
\resizebox{0.5\textwidth}{!}{%
\begin{tabular}{llrr|rr|rr}
\toprule
& & \multicolumn{2}{c}{\textbf{\tasklp}} & \multicolumn{2}{c}{\textbf{\taskke}} & \multicolumn{2}{c}{\textbf{\tasksk}} \\
\cmidrule(lr){3-4}\cmidrule(lr){5-6}\cmidrule(lr){7-8}
\textbf{Split} & \textbf{Subject} & \textbf{\#Q} & \textbf{\#A} & \textbf{\#Q} & \textbf{\#A} & \textbf{\#Q} & \textbf{\#A} \\
\midrule
\multirow{5}{*}{Train}
& Biology     & 8  & 1,148  & 8  & 1,148 & 8  & 1,148 \\
& Chemistry   & 54 & 5,921  & 52 & 5,645 & 53 & 5,755 \\
& Mathematics & 9  & 609    & NA & NA    & NA & NA \\
& Physics     & 19 & 3,103  & 18 & 2,939 & 19 & 3,073 \\
& \textbf{Total} & \textbf{90} & \textbf{10,781} & \textbf{78} & \textbf{9,732} & \textbf{80} & \textbf{9,976} \\
\midrule
\multirow{5}{*}{Test-UQ}
& Biology     & 3  & 538   & 3  & 538   & 3  & 538 \\
& Chemistry   & 12 & 1,591 & 12 & 1,591 & 12 & 1,587 \\
& Mathematics & 5  & 497   & NA & NA    & NA & NA \\
& Physics     & 2  & 470   & 2  & 469   & 2  & 470 \\
& \textbf{Total} & \textbf{22} & \textbf{3,096} & \textbf{17} & \textbf{2,598} & \textbf{17} & \textbf{2,595} \\
\midrule
\multirow{5}{*}{Test-UA}
& Biology     & 8  & 301   & 8  & 301   & 8  & 301 \\
& Chemistry   & 53 & 1,465 & 51 & 1,408 & 53 & 1,425 \\
& Mathematics & 9  & 151   & NA & NA    & NA & NA \\
& Physics     & 19 & 778   & 18 & 733   & 19 & 768 \\
& \textbf{Total} & \textbf{89} & \textbf{2,695} & \textbf{77} & \textbf{2,442} & \textbf{80} & \textbf{2,494} \\
\bottomrule
\end{tabular}
}
\caption{Split-wise question (\#Q) and answer (\#A) counts by subject and subtask.}
\label{tab:stats}
\end{table*}

\begin{figure}[ht]
    \centering
    \begin{subfigure}[b]{0.45\linewidth}
        \centering
        \includegraphics[width=\linewidth]{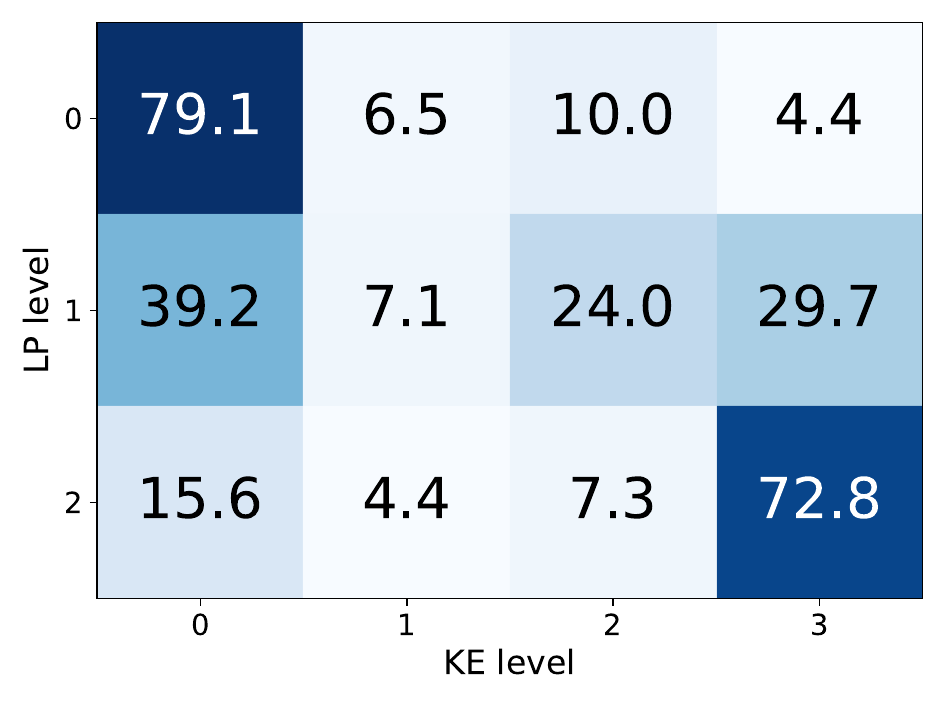}
        \caption{\tasklp--\taskke}
        \label{fig:lp-ke-heatmap}
    \end{subfigure}
    \hfill
    \begin{subfigure}[b]{0.45\linewidth}
        \centering
        \includegraphics[width=\linewidth]{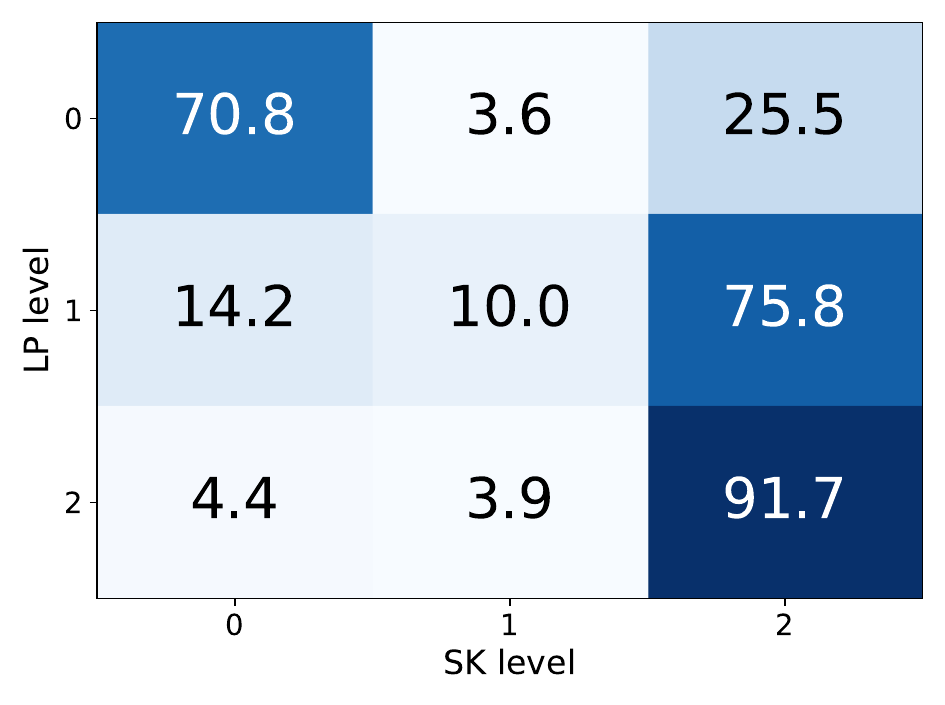}
        \caption{\tasklp--\tasksk}
        \label{fig:lp-sk-heatmap}
    \end{subfigure}
    \caption{Correlation heatmaps between levels of \tasklp and the other two subtasks.}
    \label{fig:heatmaps}
\end{figure}

We observe strong positive correlations between Learning Performance and the other two dimensions: \tasklp--\taskke ($r = 0.785$, $p < 0.001$) and \tasklp--\tasksk ($r = 0.631$, $p < 0.001$).
On the other hand, high \taskke/\tasksk scores do not necessarily translate to high \tasklp scores, and the same holds for low scores, as shown in \autoref{fig:heatmaps}.
These results validate our multidimensional framework by demonstrating that, although the dimensions are related, they capture \textit{distinct aspects} of students' understanding.
For instance, a student who fails to answer the question correctly might still demonstrate sufficient epistemic abilities and
mastery of concepts of interest. This provides more fine-grained information about students' learning progress.
\section{Modelling Approach} \label{sec4:approach}
\paragraph{LLMs as Encoders}
Alongside MLM baselines, we benchmark lightweight decoder-only LLMs as encoders, following recent work showing that autoregressive LLMs can be repurposed as strong discriminative encoders without generation~\cite{llm2vec, lookbothway, ruan2024largelanguagemodelsgoodclassifier, liu-etal-2024-rethink-llembed, lee2025nvembedimprovedtechniquestraining, qiao-etal-2025-decoderllmmaskedencoders}.
This is appealing for rubric-based ASAS, where a single input must jointly represent the student answer, the rubric item, and optionally the question and sample solution.
We therefore test whether this capacity translates into better use of rubric context and stronger generalisation to unseen questions (\autoref{sec5:results}), while avoiding the inference cost of autoregressive decoding.

\paragraph{Rubric-Retrieval}
We approach rubric-based ASAS via rubric-retrieval: during training and inference, each instance is expanded into answer--rubric pairs and encoded jointly by a language model in a cross-encoder setup.
MLMs~\cite{devlin2019bertpretrainingdeepbidirectional} separate segments with the \texttt{[SEP]} token,
while LLM-based encoders use the best-performing \textit{structured} format from \citet{ruan2024largelanguagemodelsgoodclassifier}.
For detailed input formatting, see \autoref{appendix:input-formatting}.

The alternative to a cross-encoder is a \textit{bi-encoder} (Siamese) setup, which encodes answers and rubric items separately and
compares them with cosine similarity or dot product.
Both architectures are established in the sentence embedding and retrieval literature~\cite{reimers2019sentencebertsentenceembeddingsusing, muennighoff2023mtebmassivetextembedding};
bi-encoders are favoured there primarily for scalability, since embeddings can be precomputed and reused across large candidate pools,
whereas cross-encoders must re-encode every pair but generally achieve higher pairwise accuracy.
We adopt the cross-encoder because each instance has only a small rubric candidate set, so the additional encoding cost is negligible,
while joint encoding enables direct cross-attention between answer and rubric segments,
yielding richer interaction features for alignment.
This formulation is convenient for settings where rubric sets vary across questions, unlike sequence-classification setups that assume a fixed label space.

Given instance $i$ and rubric item $r_j^i \in R^i$, the encoder $f_\theta$ produces a scalar alignment score:
\[
z_{i,j} = f_\theta(q_i, a_i, s_i, r_j^i).
\]
Here, $z_{i,j} \in \mathbb{R}$ indicates how well rubric item $r_j^i$ matches answer $a_i$,
optionally conditioned on question $q_i$ and sample solution $s_i$.

Related ASAS work has explored pairwise or contrastive supervision, for example by comparing student answers
to candidate answers from each level via similarity-based scoring~\cite{bexte-etal-2022-similarity} or
by framing answer--rubric alignment as an NLI-style decision~\cite{sonkar2024automatedlonganswergrading}.
However, because the rubrics in \textsc{Alice} can be semantically overlapping rather than mutually exclusive (\autoref{subsection:rubentail}) and are also not additive, unlike in \citet{sonkar2024automatedlonganswergrading},
binary matching alone is insufficient:
the model must explicitly \textit{contrast} the correct rubric against alternatives within the same set.
We therefore use softmax cross-entropy over the rubric set. When a training batch contains instances with different numbers of rubric levels, we mask
additional alignment scores to accommodate variable rubric-set sizes.

This objective directly models competition among rubric levels and yields normalised probabilities over all rubric candidates.
As ablations, we also evaluate contrastive loss and a bi-encoder variant, similar to previous work on \alicelp in \autoref{appendix:lp-ablation}.

The probability of rubric item $j$ is defined as
\begin{equation}
P_\theta(j \mid i) =
\frac{\exp(z_{i,j})}
{\sum_{k=1}^{|R^i|} \exp(z_{i,k})}.
\label{eq:sce_softmax_active}
\end{equation}

Let $y_i$ denote the gold rubric index for instance $i$. The loss is
\begin{equation}
\ell^{(i)}_{\text{SCE}} = - \log P_\theta(y_i \mid i).
\label{eq:sce_loss_active}
\end{equation}

Softmax normalisation and prediction are computed only over the rubric candidates associated with the same question or \taskke/\tasksk item. At inference time, we predict
\[
\hat{y}_i = \arg\max_{j \in \{1,\dots, |R^i|\}} P_\theta(j \mid i).
\]
Masked candidates are excluded before normalisation.

\paragraph{Sequence Classification}
We also benchmark \textsc{Alice} with a modelling approach commonly used in prior ASAS work~\cite{domain-pretrain-asas, asas-transformer}.
Specifically, the student answer and the corresponding sample solution are concatenated and provided as input to a language model
with a classification head on top. We refer to this setup as the \textit{sequence-classification} approach.

This baseline is applied to all \textsc{Alice} subtasks.
For \aliceke and \alicesk, some questions do not include all intermediate rubric levels.
Because the sequence-classification head assumes a fixed, uniform set of level labels across questions, we map levels to indices according to the global level scheme
(\textit{No use} $\rightarrow$ 0, \dots, \textit{Targeted use} $\rightarrow$ 3) and simply omit indices for levels that are not defined for a given question, rather than re-indexing the remaining levels.
Note that because scoring in these two subtasks depends on \taskke and \tasksk, we prepend the corresponding \taskke/\tasksk name to the model input.

\paragraph{Zero-shot Prompting}

In addition to the fine-tuning approach, we run zero-shot prompting with both open-weight and API-hosted proprietary large language models to perform rubric-based scoring
by providing the same input components used for encoder-based models, including the student response and the corresponding
rubric items, optionally augmented with the question prompt and sample solution.
The models are instructed to select the most appropriate rubric level for each instance.

\paragraph{Input Format.}
We compare base vs. full-context input configurations. For rubric-retrieval approaches,
the essential inputs are the \textit{answer} and \textit{rubrics}; the \textit{question} and \textit{sample solution} are optional context.
For sequence classification, the \textit{question} and \textit{rubrics} are optional context.
For zero-shot LLM prompting, we experiment with three input formats: (i) full context plus rubrics, (ii) no context plus rubrics, and (iii) no context with label names only.
See \autoref{appendix:input-formatting} and \autoref{appendix:llms} for input format examples and prompts.
\section{Benchmarking on \textsc{Alice}} \label{sec5:results}

\begin{table*}[t]
\centering
\small
\setlength{\tabcolsep}{3pt}

\begin{subtable}[t]{0.49\textwidth}
\centering
\caption{UA Macro-F1}
\resizebox{\linewidth}{!}{%
\begin{tabular}{llcc|cc|cc}
\toprule
Approach & Model & \multicolumn{2}{c}{\alicelp} & \multicolumn{2}{c}{\aliceke} & \multicolumn{2}{c}{\alicesk} \\
\cmidrule(lr){3-4}\cmidrule(lr){5-6}\cmidrule(lr){7-8}
Rub. Ret. & Input format & \textit{ar} & \textit{+qs} & \textit{ar} & \textit{+qs} & \textit{ar} & \textit{+qs} \\
\midrule
 & XLM-RoBERTa-Long & 67.3 & \cellcolor{red!15}64.5 & 65.7 & \cellcolor{red!15}58.9 & 65.5 & \cellcolor{red!15}62.9 \\
 & mmBERT & 71.7 & \cellcolor{red!15}69.0 & 69.5 & 70.1 & 70.3 & \cellcolor{red!15}68.6 \\
 & Llama-3.2-1B & 72.1 & 73.3 & 70.8 & \cellcolor{red!15}70.5 & 69.4 & 69.5 \\
 & Llama-3.2-1B-Instruct & 72.3 & 72.9 & 70.3 & \textbf{72.5} & 70.0 & 70.0 \\
 & Llama-3.2-3B & 73.8 & 75.4 & 72.3 & 72.4 & 69.9 & 70.0 \\
 & Llama-3.2-3B-Instruct & 74.8 & \textbf{76.7} & 72.4 & \cellcolor{red!15}71.0 & 70.6 & \textbf{71.8} \\
\midrule
Seq. Class. & Input format & \textit{sa} & \textit{+qr} & \textit{sa} & \textit{+qr} & \textit{sa} & \textit{+qr} \\
\midrule
 & XLM-RoBERTa-Long & \underline{68.0} & \cellcolor{red!15}\underline{65.4} & 64.5 & \underline{64.8} & \underline{66.5} & \underline{66.7} \\
 & mmBERT & \underline{72.1} & \cellcolor{red!15}\underline{71.0} & 55.2 & 68.9 & 68.0 & \underline{70.8} \\
 & Llama-3.2-1B & \underline{73.5} & \cellcolor{red!15}72.7 & 70.0 & \cellcolor{red!15}68.8 & 67.3 & 68.6 \\
 & Llama-3.2-1B-Instruct & \underline{75.3} & \cellcolor{red!15}\underline{73.6} & 54.8 & 69.0 & \underline{70.1} & \cellcolor{red!15}69.0 \\
 & Llama-3.2-3B & \underline{75.2} & \cellcolor{red!15}74.9 & 54.7 & 71.4 & 69.9 & \underline{70.1} \\
 & Llama-3.2-3B-Instruct & \underline{76.2} & \cellcolor{red!15}75.1 & 54.7 & \underline{72.0} & 69.2 & 71.7 \\
\bottomrule
\end{tabular}
}
\end{subtable}
\hfill
\begin{subtable}[t]{0.49\textwidth}
\centering
\caption{UQ Macro-F1}
\resizebox{\linewidth}{!}{%
\begin{tabular}{llcc|cc|cc}
\toprule
Approach & Model & \multicolumn{2}{c}{\alicelp} & \multicolumn{2}{c}{\aliceke} & \multicolumn{2}{c}{\alicesk} \\
\cmidrule(lr){3-4}\cmidrule(lr){5-6}\cmidrule(lr){7-8}
Rub. Ret. & Input format & \textit{ar} & \textit{+qs} & \textit{ar} & \textit{+qs} & \textit{ar} & \textit{+qs} \\
\midrule
 & XLM-RoBERTa-Long & 62.2 & \cellcolor{red!15}55.8 & 50.9 & \cellcolor{red!15}38.3 & 60.1 & \cellcolor{red!15}46.6 \\
 & mmBERT & 61.5 & \cellcolor{red!15}60.0 & 52.6 & 53.7 & 53.9 & 56.2 \\
 & Llama-3.2-1B & 64.9 & 65.9 & 54.4 & 56.2 & 54.7 & 61.8 \\
 & Llama-3.2-1B-Instruct & 66.2 & \cellcolor{red!15}65.5 & 55.6 & \cellcolor{red!15}54.3 & 60.2 & \cellcolor{red!15}56.6 \\
 & Llama-3.2-3B & 64.7 & 66.8 & 53.6 & \textbf{60.0} & 57.6 & \textbf{65.4} \\
 & Llama-3.2-3B-Instruct & 62.6 & \textbf{69.5} & 57.4 & 57.8 & 56.5 & 65.0 \\
\midrule
Seq. Class. & Input format & \textit{sa} & \textit{+qr} & \textit{sa} & \textit{+qr} & \textit{sa} & \textit{+qr} \\
\midrule
 & XLM-RoBERTa-Long & 58.1 & \cellcolor{red!15}\underline{57.7} & 44.6 & \cellcolor{red!15}\underline{42.6} & 52.0 & \underline{60.6} \\
 & mmBERT & \underline{62.2} & \cellcolor{red!15}\underline{61.6} & 43.4 & 53.4 & 45.9 & 47.1 \\
 & Llama-3.2-1B & 63.1 & \cellcolor{red!15}61.7 & 49.3 & \cellcolor{red!15}48.6 & 44.6 & 51.0 \\
 & Llama-3.2-1B-Instruct & 63.3 & 63.8 & 40.5 & 50.9 & 47.0 & 53.8 \\
 & Llama-3.2-3B & 64.5 & 64.8 & 40.4 & 51.9 & 53.3 & 60.6 \\
 & Llama-3.2-3B-Instruct & \underline{66.4} & \cellcolor{red!15}65.1 & 41.0 & 52.7 & 54.2 & 59.5 \\
\bottomrule
\end{tabular}
}
\end{subtable}

\vspace{0.5em}

\begin{subtable}[t]{0.74\textwidth}
\centering
\caption{Zero-shot UQ Macro-F1}
\scriptsize
\setlength{\tabcolsep}{3pt}
\begin{tabular}{lccc|ccc|ccc}
\toprule
Model & \multicolumn{3}{c}{\tasklp} & \multicolumn{3}{c}{\taskke} & \multicolumn{3}{c}{\tasksk} \\
\cmidrule(lr){2-4}\cmidrule(lr){5-7}\cmidrule(lr){8-10}
 & \textit{-r} & \textit{+r} & \textit{+qs+r} & \textit{-r} & \textit{+r} & \textit{+qs+r} & \textit{-r} & \textit{+r} & \textit{+qs+r} \\
\midrule
Mistral-7B-Instruct & 41.1 & 48.4 & 47.5 & 38.8 & 37.3 & \cellcolor{red!15}38.5 & 28.8 & 29.4 & 38.6 \\
Llama-3.1-8B-Instruct & \textbf{45.7} & 45.9 & \cellcolor{red!15}45.1 & 31.9 & 33.9 & \cellcolor{red!15}31.2 & 48.8 & 36.8 & \cellcolor{red!15}31.9 \\
Llama-3.3-70B-Instruct & 42.0 & 60.6 & 58.2 & 40.4 & 42.0 & \cellcolor{red!15}36.6 & 34.4 & 25.5 & \cellcolor{red!15}26.4 \\
GPT-4o-mini & 36.0 & 60.2 & 60.6 & 37.1 & 46.1 & \cellcolor{red!15}36.9 & \textbf{54.1} & \textbf{58.0} & \textbf{60.3} \\
GPT-5-mini & 43.3 & \textbf{64.0} & \textbf{66.1} & \textbf{42.6} & \textbf{59.8} & \textbf{59.9} & 43.9 & 57.6 & 59.5 \\
\bottomrule
\end{tabular}
\end{subtable}

\caption{Main Macro-F1 results on \textsc{Alice}. Upper panels show fine-tuned UA and UQ results; the lower panel shows zero-shot UQ prompting. In the upper panels, bold marks the best result per subtask; in the lower panel, it marks the best result per input format and subtask. The \underline{underline} in the upper panels marks cells where sequence classification outperforms rubric-retrieval. \colorbox{red!15}{Red cells} mark full-context results (\textit{+qs}/\textit{+qr}/\textit{+qs+r}) that underperform the base input format (\textit{ar}/\textit{sa}/\textit{-r}) for that model and subtask.}
\label{tab:main-results}
\end{table*}

\subsection{Experimental Setup}
We evaluate rubric-retrieval, sequence classification, and zero-shot LLM prompting across all three \textsc{Alice} subtasks (\alicelp, \aliceke, \alicesk) on the split in \autoref{sec:3}.
Test-UA measures generalisation to unseen student answers for seen questions, while Test-UQ measures generalisation to entirely unseen questions, the harder setting.
For fine-tuning, we compare two MLM baselines against four lightweight LLM encoders, with and without full question and sample-solution context.
Zero-shot prompting evaluates whether rubric text alone is sufficient to guide large models without task-specific training.
We focus on macro-F1 and UQ generalisation.
Complementary QWK results are reported in \autoref{tab:qwk-results}.
Model configurations and hyperparameters are provided in \autoref{appendix:hyperparameters}.
We further apply rubric-retrieval to \textsc{ASAP-SAS}; see \autoref{appendix:asap-sas}.
\subsection{Results and Discussion}

\begin{figure*}[ht]
    \centering
    \includegraphics[width=0.82\linewidth]{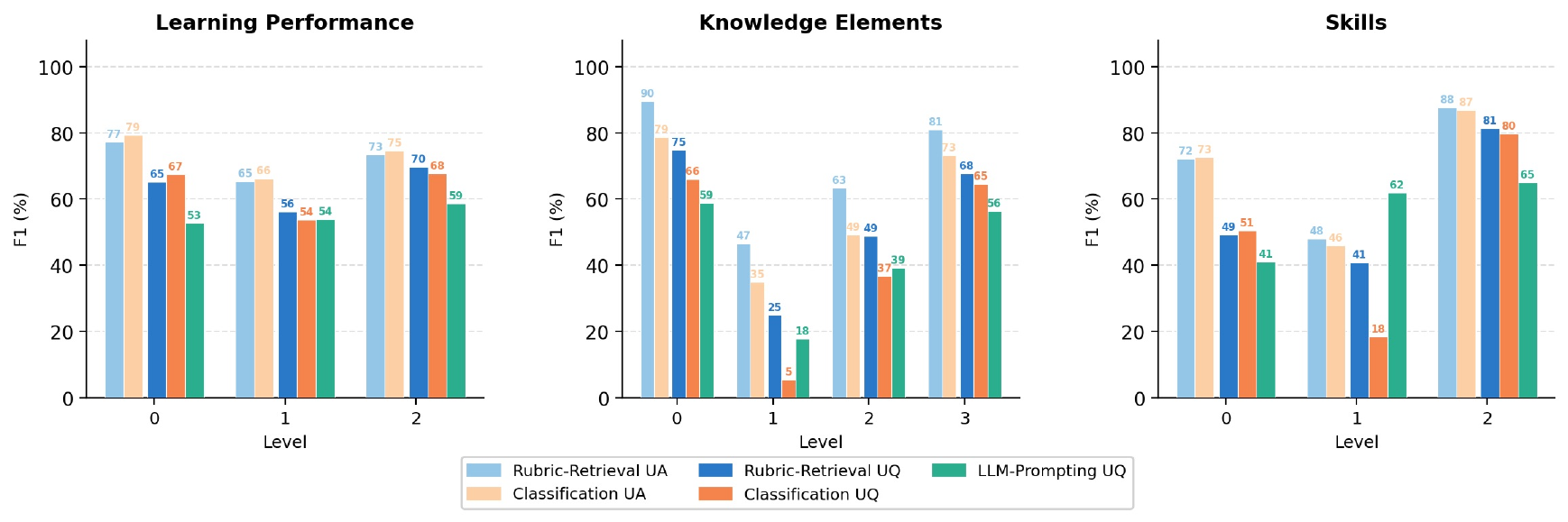}
    \caption{Per-level F1 across \textsc{Alice} subtasks and evaluation settings.}
    \label{fig:per-level-f1}
\end{figure*}

\paragraph{RQ1: What are the differences between rubric-retrieval and the classification baseline, and how well do the fine-tuned approaches generalise to unseen questions?}

Rubric-retrieval outperforms sequence classification more decisively on UQ than on UA.
On UQ, it shows consistent gains across all three subtasks, with larger margins on \aliceke and \alicesk than on \alicelp.
On UA, the gap narrows substantially; sequence classification remains competitive or slightly ahead on \alicelp.
For \alicelp, sequence classification remains competitive as overall learning performance can often be approximated by comparing a student answer with the sample solution under fixed score labels across questions; accordingly, both approaches degrade similarly on Test-UQ for this subtask.
The advantage of rubric-retrieval is clearer for \aliceke and \alicesk for two reasons.
First, as shown in \autoref{fig:heatmaps}, \taskke and \tasksk scores are not always correlated with \tasklp, indicating that scoring these dimensions requires attending to the rubric descriptions rather than relying primarily on the sample solution.
Second, rubric-retrieval may be less sensitive to label imbalance: by aligning responses with rubric candidates rather than predicting from a fixed class distribution, the model is not tied as directly to global class frequencies.
For \aliceke and \alicesk, sequence classification without rubric context degrades more severely on Test-UQ; rubric-retrieval is more robust by aligning to rubric text rather than fixed labels.
Adding rubric context to sequence classification (\textit{+qr}) substantially narrows this gap, confirming that rubric information, rather than training labels alone, is the key driver of question-level generalisation.
\autoref{fig:per-level-f1} shows the same pattern at the level granularity: intermediate levels remain difficult, but rubric-retrieval is more stable for the fine-grained knowledge-element and skill distinctions.

\paragraph{RQ2: Are LLMs as encoders suitable for ASAS?}
The results support lightweight LLMs as strong ASAS encoders.
Under rubric-retrieval, LLM encoders produce the best results for all subtasks: \textit{Llama-3.2-3B-Instruct} is strongest on \alicelp UQ, while \textit{Llama-3.2-3B} gives the best UQ results on \aliceke and \alicesk.
MLM baselines remain competitive in some fixed-label sequence-classification settings, such as \aliceke UQ with \textit{mmBERT}, but these exceptions do not overturn the broader trend that larger, decoder-based encoders tend to generalise better to unseen questions on \textsc{Alice}.

Zero-shot prompting with strong proprietary models is also competitive: \textit{GPT-5-mini} approaches fine-tuned performance on \alicelp and \aliceke, while \textit{GPT-4o-mini} is close to the sequence-classification baseline on \alicesk.
Among open-weight models, \textit{Llama-3.3-70B-Instruct} substantially outperforms the smaller open-weight models on \alicelp and approaches \textit{GPT-4o-mini} on that subtask; however, it underperforms the proprietary models on \aliceke and \alicesk, suggesting that, on \textsc{Alice}, scale alone does not close the gap on the more fine-grained subtasks.
Overall, the results support LLMs both as encoders and, in stronger proprietary settings, as zero-shot scorers; supervised rubric-retrieval remains the most reliable configuration.

\paragraph{RQ3: How does additional context affect model performance?}

Additional context helps, but the benefit varies across subtasks and approaches.
For rubric-retrieval, the full \textit{+qs} format benefits \aliceke and \alicesk more consistently than \alicelp, particularly with stronger LLM encoders.
For sequence classification, \alicelp benefits less from added question and rubric information, whereas \aliceke and \alicesk gain more substantially, consistent with the greater rubric-dependence of these subtasks.
In zero-shot prompting, rubric text is the key driver of improvement: moving from label names to rubric descriptions often yields substantial gains, especially for stronger proprietary models.
Smaller open-weight models benefit less, and for \aliceke and \alicesk, extending the prompt beyond rubric text to include the full question context can even degrade performance, suggesting these models struggle to selectively leverage additional information.
On \textsc{Alice}, rubric text thus emerges as the most informative contextual signal; full context provides additional benefit mainly when the model can leverage it reliably.

\section{Conclusion}
We introduce \textsc{Alice}, a German rubric-based ASAS benchmark for learning performance,
knowledge elements, and skills.
Beyond the dataset, we provide a rubric-retrieval benchmark formulation that accommodates question-specific
rubric structures and serves as a practical alternative to fixed-label sequence classification.

Our experiments show that rubric-retrieval and sequence classification are both competitive on \alicelp, while rubric-retrieval shows clearer advantages on \aliceke and \alicesk.
The results further suggest that for the fine-tuned models, rubric information is particularly important for these latter two subtasks, whereas for \alicelp the relative benefit over sample-solution-focused scoring is smaller. For
zero-shot LLM prompting, per-level rubric text is important across all subtasks.

\section{Limitations}
This work has several limitations. First, \textsc{Alice} focuses on German-language scientific education, and while the dataset design is general, results may not directly transfer to other languages or subject domains. Second, although we evaluate a range of encoder architectures, we restrict our study to discriminative models and do not explore generative scoring or feedback generation. Future work on feedback generation and pedagogical capabilities of LLMs could benefit from integrating fine-grained assessment dimensions such as knowledge elements and skills.

While rubric-retrieval proves effective and generalises well to unseen questions and question-dependent rubric structures, it requires expanding each answer into multiple answer--rubric pairs, which increases training and inference costs and limits scalability to larger models. Finally, rubric-based scoring relies on the availability of high-quality rubrics, which may not always be accessible in real-world settings. Exploring LLM-augmented rubrics and other contextual information is a promising direction for future research.

\section{Ethical Statement}
The \textsc{Alice} benchmark is constructed from student responses to assessment items collected in routine
classroom settings. We obtained all data under appropriate educational governance and privacy protections; identifiers were removed,
and responses were anonymised before annotation to safeguard student privacy.
Annotators were compensated for their work,
and annotation quality was monitored through pilot rounds, rubric revisions, and adjudication.

We acknowledge that automated short-answer scoring systems can influence educational outcomes, and misuse may unfairly advantage or disadvantage
students if deployed without appropriate safeguards.
\textsc{Alice} is intended for research on rubric-based short-answer scoring.
Models evaluated on this benchmark should be interpreted as tools for supporting instructional practice,
not as replacements for expert human judgment. We encourage researchers to consider bias, fairness,
and interpretability when developing and reporting models on \textsc{Alice}, and
to avoid applications that could reinforce inequitable treatment of learners.

\section*{Acknowledgments}

This work was supported by the Volkswagen Foundation through the project \emph{From Machine Learning to Machine Teaching: Making Machines AND Humans Smarter} (ML2MT).
The GPU resources used in this work were provided by Hessian.ai.

\bibliography{ref}
\appendix

\section{Dataset Example} \label{appendix:instance}
See \autoref{fig:dataset_instance}. In this example, level 1 is absent from the knowledge-element rubrics because it is reserved for the ``Use without content'' scenario and is therefore not defined for this Physics item. Likewise, the skills rubric for this question does not include level 1 (``partially present'').
\begin{figure*}[ht]
\centering
\newcommand{\cardwidth}{0.92\linewidth}

\definecolor{selgreen}{RGB}{210,245,210}

\tcbset{
  card/.style={
    enhanced,
    width=\cardwidth,
    colback=white,
    colframe=black!60,
    boxrule=0.5pt,
    arc=2mm,
    left=2mm,
    right=2mm,
    top=1.5mm,
    bottom=1.5mm
  },
  header/.style={
    colback=black!10,
    fontupper=\bfseries,
    left=1mm,
    right=1mm
  }
}

\begin{tcolorbox}[card, title=Question Context Information]
\textbf{Prompt:} Based on the patterns and measured values, derive two ``the more... the more''
statements for the alignment of the PV cell.

\vspace{1mm}

\textbf{Sample Solution:}\\
The more perpendicular the PV cell is aligned to the radiation source, the higher the electrical
energy. The more light reaches the PV cell (the less the PV cell is shaded), the higher the
electrical energy.
\end{tcolorbox}

\vspace{1.5mm}

\begin{tcolorbox}[card, title=Student Response]
\textbf{Answer:} The more directly the light source shines on the solar panel, the higher the
generated voltage. The more of the panel was covered, the less voltage was generated.
\end{tcolorbox}

\vspace{1.5mm}

\begin{tcolorbox}[card, title=Learning Performance]
\textbf{Level 0:} The student does not formulate any ``the more... the more'' statement about the angle of incidence or the illuminated area.\\[1mm]
\textbf{Level 1:} The student formulates one statement about either the angle or the illuminated area.\\[1mm]
\colorbox{selgreen}{\parbox{\linewidth}{\textbf{Level 2:} The student formulates two statements, covering both the angle and the illuminated area.}}
\end{tcolorbox}

\vspace{1.5mm}

\begin{tcolorbox}[card, title=Knowledge Elements]

\textbf{Electrical Energy}\\
Level 0: The term or similar meanings are not mentioned.\\
Level 2: Non-targeted use: The students use the voltage to explain something other than energy conversion.\\
\colorbox{selgreen}{\parbox{\linewidth}{\textbf{Level 3:} Targeted use: voltage is used to make statements about energy conversion.}}

\vspace{1mm}

\textbf{Radiation Energy}\\
Level 0: The term or similar meanings are not mentioned.\\
Level 2: Non-targeted use: The students use angles or shading to explain something other than energy conversion.\\
\colorbox{selgreen}{\parbox{\linewidth}{\textbf{Level 3:} Targeted use: angle or shading is used to make statements about energy conversion.}}

\vspace{1mm}

\textbf{Conversion}\\
Level 0: The term or similar meanings are not mentioned.\\
Level 2: Non-targeted use: The students use transformation concepts to explain something other than the conversion of radiant energy into electrical energy.\\
\colorbox{selgreen}{\parbox{\linewidth}{\textbf{Level 3:} Targeted use: The student indicates that radiation is converted into electrical energy.}}
\end{tcolorbox}

\vspace{1.5mm}

\begin{tcolorbox}[card, title=Skills]
\vspace{1mm}
\textbf{Reasoning} \\
Level 0: Not present: No explanation based on the experimental results.\\[1mm]
\colorbox{selgreen}{\parbox{\linewidth}{\textbf{Level 2:} Present: Explains patterns using the experimental results and includes at least one ``the more... the more'' statement.}}
\end{tcolorbox}
\caption{Illustrative benchmark instance, translated from German to English. Selected rubric levels are highlighted in green.}
\label{fig:dataset_instance}
\end{figure*}
\FloatBarrier

\section{Input Formats of LLMs} \label{appendix:input-formatting}
\citet{ruan2024largelanguagemodelsgoodclassifier} compare two input formats for LLM-based classification: \textit{natural language} and \textit{structured}. The former uses natural-language sequence boundaries, while the latter marks sequence boundaries with XML-style tags. The \textit{structured} formatting approach achieved better results, so we use it in our experiments.

\paragraph{Sequence Classification}\mbox{}\par
\begin{lstlisting}[
  basicstyle=\ttfamily\small,
  breaklines=true,
  breakatwhitespace=false,
  columns=fullflexible,
  keepspaces=true
]
<Frage>Leitet auf Grundlage der Muster und der Messwerte zwei "Je - desto" Aussagen fuer die Ausrichtung der PVzelle ab.</Frage>
<Loesung>Je senkrechter die PVzelle zur Strahlungsquelle ausgerichtet ist, desto hoeher ist die Elektrische Energie. Je mehr Licht die PVzelle erreicht (je geringer die PVzelle verschattet ist), desto hoeher ist die Elektrische Energie.</Loesung>
<Antwort>Je anders der Winkel ist desto anders sind auch die Messwerte</Antwort>
<Rubrik>nicht richtig:Die SuS formulieren keine je-desto-Aussagen zum Einstrahlungswinkel und beschienenen Flaeche.</Rubrik>
<Rubrik>teilweise richtig:Die SuS formulieren eine je-desto-Aussagen zu Einstrahlungswinkel oder beschienenen Flaeche.</Rubrik>
<Rubrik>richtig:Die SuS formulieren zwei je-desto-Aussagen zu Einstrahlungswinkel und beschienenen Flaeche.</Rubrik>
\end{lstlisting}

\paragraph{Rubric-Retrieval}\mbox{}\par

\begin{lstlisting}[
  basicstyle=\ttfamily\small,
  breaklines=true,
  breakatwhitespace=false,
  columns=fullflexible,
  keepspaces=true
]
# candidate rubric_level=0, label=0
<Frage>Leitet auf Grundlage der Muster und der Messwerte zwei "Je - desto" Aussagen fuer die Ausrichtung der PVzelle ab.</Frage>
<Loesung>Je senkrechter die PVzelle zur Strahlungsquelle ausgerichtet ist, desto hoeher ist die Elektrische Energie. Je mehr Licht die PVzelle erreicht (je geringer die PVzelle verschattet ist), desto hoeher ist die Elektrische Energie.</Loesung>
<Antwort>Je anders der Winkel ist desto anders sind auch die Messwerte</Antwort>
<Rubrik>nicht richtig:Die SuS formulieren keine je-desto-Aussagen zum Einstrahlungswinkel und beschienenen Flaeche.</Rubrik>

# candidate rubric_level=1, label=1
<Frage>Leitet auf Grundlage der Muster und der Messwerte zwei "Je - desto" Aussagen fuer die Ausrichtung der PVzelle ab.</Frage>
<Loesung>Je senkrechter die PVzelle zur Strahlungsquelle ausgerichtet ist, desto hoeher ist die Elektrische Energie. Je mehr Licht die PVzelle erreicht (je geringer die PVzelle verschattet ist), desto hoeher ist die Elektrische Energie.</Loesung>
<Antwort>Je anders der Winkel ist desto anders sind auch die Messwerte</Antwort>
<Rubrik>teilweise richtig:Die SuS formulieren eine je-desto-Aussagen zu Einstrahlungswinkel oder beschienenen Flaeche.</Rubrik>

# candidate rubric_level=2, label=0
<Frage>Leitet auf Grundlage der Muster und der Messwerte zwei "Je - desto" Aussagen fuer die Ausrichtung der PVzelle ab.</Frage>
<Loesung>Je senkrechter die PVzelle zur Strahlungsquelle ausgerichtet ist, desto hoeher ist die Elektrische Energie. Je mehr Licht die PVzelle erreicht (je geringer die PVzelle verschattet ist), desto hoeher ist die Elektrische Energie.</Loesung>
<Antwort>Je anders der Winkel ist desto anders sind auch die Messwerte</Antwort>
<Rubrik>richtig:Die SuS formulieren zwei je-desto-Aussagen zu Einstrahlungswinkel und beschienenen Flaeche.</Rubrik>
\end{lstlisting}

\section{Full \taskke and \tasksk Items List} \label{appendix:full-item-lists}

\begin{table*}[ht]
\centering
\scriptsize
\caption{%
  Knowledge Element (\taskke) item names per subject.
  Each entry is a domain concept that students are scored on independently.
}
\label{tab:ke-names}
\begin{tabular}{lp{10cm}}
\toprule
\textbf{Subject} & \textbf{\taskke items} \\
\midrule
Biology & Gendrift (Genetic Drift), Genetik (Genetics), Mutation (Mutation), Natürliche Selektion (Natural Selection), Sexuelle Selektion (Sexual Selection), Variation (Variation), Wahrscheinlichkeit (Probability), Zufall (Randomness) \\
\midrule
Chemistry & Abklingfunktion (Decay Function), Aggregatzustand (State of Matter), Aggregatzustände (States of Matter), Aktivierungsenergie (Activation Energy), Anzahl an Gasteilchen (Number of Gas Particles), Dissoziation (Dissociation), Druck (Pressure), Druckänderung (Pressure Change), Dynamisches Gleichgewicht (Dynamic Equilibrium), Eduktregenerierung (Reactant Regeneration), Einflussfaktor Druck (Influencing Factor: Pressure), Einflussfaktor Temperatur (Influencing Factor: Temperature), Endotherme Rückreaktion (Endothermic Reverse Reaction), Energieumsatz (Energy Conversion), Exemplarität/Deduktion (Exemplarity/Deduction), Exotherme Hinreaktion (Exothermic Forward Reaction), Gasentwicklung (Gas Evolution), Geschlossenes System (Closed System), Gleichgewichtseinstellung (Equilibrium Establishment), Gleichgewichtskonstante (Equilibrium Constant), Gleichzeitigkeit (Simultaneity), Hinreaktion (Forward Reaction), Inaktivierung (Inactivation), Katalysator (Catalyst), Katalysatorrückbildung (Catalyst Regeneration), Kollision (Collision), Konzentration (Concentration), Kostenintensivität (Cost Intensity), Ladungsträger (Charge Carrier), Leitfähigkeit (Conductivity), Makroskopisches Reaktionsende (Macroscopic Reaction Endpoint), Masse (Mass), Mindestenergie (Minimum Energy), Mindestenergie/Aktivierungsenergie (Minimum Energy/Activation Energy), Mittelwert (Mean), Neueinstellung (Re-Establishment of Equilibrium), Offenes System (Open System), Produktentfernung (Product Removal), Reaktionsenthalpie (Reaction Enthalpy), Reaktionsgeschwindigkeit (Reaction Rate), Reaktionsrate (Reaction Rate), Reaktionsweg (Reaction Pathway), Reaktionszeit (Reaction Time), Rückreaktion (Reverse Reaction), Sekantensteigung (Secant Slope), Stoffmenge (Amount of Substance), Stoffumsatz (Substance Conversion), Stoffumwandlung (Substance Transformation), Stoßwahrscheinlichkeit (Collision Probability), Stoßwirksamkeit (Collision Effectiveness), System (System), Sättigungskurve (Saturation Curve), Sättigungsfunktion (Saturation Function), Tangentensteigung (Tangent Slope), Teilchengeschwindigkeit (Particle Velocity), Teilcheninteraktion (Particle Interaction), Temperatur (Temperature), Ungleichheit von Entitäten (Inequality of Entities), Unvollständigkeit (Incompleteness), Volumen (Volume), Wiedereinstellung (Re-Establishment), Wirksamkeit Katalysator (Catalyst Effectiveness), Zeitintervall (Time Interval), Zeitunabhängigkeit (Time Independence), Zellgift (Cytotoxin), Zellweger-Syndrom (Zellweger Syndrome), Zerteilungsgrad (Degree of Dispersion), Zufallsfehler (Random Error) \\
\midrule
Mathematics & -- \\
\midrule
Physics & Chemische Energie (Chemical Energy), Elektrische Energie (Electrical Energy), Energieversorgung (Gesellschaft) (Energy Supply: Society), Energieversorgung (Ökologie) (Energy Supply: Ecology), Kinetische Energie (Kinetic Energy), Strahlungsenergie (Radiation Energy), Thermische Energie (Thermal Energy), Umwandlung (Conversion) \\
\bottomrule
\end{tabular}
\end{table*}

\begin{table*}[ht]
\centering
\scriptsize
\begin{tabular}{lp{13cm}}
\toprule
\textbf{Subject} & \textbf{\tasksk items} \\
\midrule
Biology & Beschreiben den Experimentaufbau (Describe the Experimental Setup), Bewerten der Daten (Evaluate the Data), Claim (Claim), Erläutern die Erklärungskraft eines Modells (Explain the Explanatory Power of a Model), Evidence (Evidence), Formulieren eine Forschungsfrage (Formulate a Research Question), Formulieren eine Hypothese (Formulate a Hypothesis), Forschungsfragen stellen (Pose Research Questions), Hypothesen/Vermutungen aufstellen (Formulate Hypotheses/Assumptions), Konzentration auf bestimmte Aspekte (Focus on Specific Aspects), Reasoning (Reasoning) \\
\midrule
Chemistry & Analysieren Zusammenhänge, die im Modell expliziert werden (Analyze Relationships Made Explicit in the Model), Auseinandersetzung mit dem Phänomen (Engagement with the Phenomenon), Beobachten und Messen (Observing and Measuring), Beschreiben auf Basis eines Phänomens relevante Variablen, Systeme und/oder Konzepte (Describe Relevant Variables, Systems, and/or Concepts Based on a Phenomenon), Beschreiben den Experimentieraufbau (Describe the Experimental Setup), Beschreiben die vorliegenden Daten (Describe the Available Data), Bestimmen mithilfe der Daten relevante Aspekte / Werte (durch Berechnung) (Determine Relevant Aspects/Values from Data (by Calculation)), Beziehen die relevanten Aspekte / Werte auf die vorhandenen Informationen (Relate the Relevant Aspects/Values to the Available Information), Claim (Claim), Erklären, warum die Daten die Behauptung stützen oder widerlegen (Explain Why the Data Support or Refute the Claim), Erläutern Rückschlüsse zu den Implikationen ihrer Ergebnisse (Explain Inferences about the Implications of Their Results), Erläutern die Erklärungskraft eines Modells (Explain the Explanatory Power of a Model), Evaluieren Stärken und Schwächen der eigenen Modellierung durch Vergleiche mit Konsensmodellen (Evaluate Strengths and Weaknesses of Own Modelling through Comparison with Consensus Models), Fassen Erkenntnisse der Modellierung zusammen (Summarize Insights from Modelling), Fassen die Daten zusammen, um die Frage zu beantworten (Summarize the Data to Answer the Question), Formulieren eine Hypothese (Formulate a Hypothesis), Identifizieren die relevanten Daten oder den Beleg, der die Behauptung belegt (Identify the Relevant Data or Evidence Supporting the Claim), Konzentration auf bestimmte Aspekte (Focus on Specific Aspects), Nutzen die Manipulation zur Erklärung des Phänomens (Use Manipulation to Explain the Phenomenon), Reasoning (Reasoning), Stellen Veränderungen der Modellkomponenten angemessen dar (Represent Changes in Model Components Appropriately), Verknüpfen im Modell auftretende Variablen und formulieren Hypothesen zu deren Zusammenhang (Link Variables in the Model and Formulate Hypotheses about Their Relationship), Wählen einen geeigneten Versuchsansatz aus (Select an Appropriate Experimental Approach) \\
\midrule
Mathematics & -- \\
\midrule
Physics & Beschreiben den Experimentieraufbau (Describe the Experimental Setup), Beschreiben die vorliegenden Daten (Describe the Available Data), Bestimmen mithilfe der Daten relevante Aspekte / Werte (durch Berechnung) (Determine Relevant Aspects/Values from Data (by Calculation)), Beziehen die relevanten Aspekte / Werte auf die vorhandenen Informationen (Relate the Relevant Aspects/Values to the Available Information), Claim (Claim), Fassen Erkenntnisse der Modellierung zusammen (Summarize Insights from Modelling), Fassen die Daten zusammen, um die Frage zu beantworten (Summarize the Data to Answer the Question), Formulieren erste Zusammenhänge, Muster und Strukturen in den Daten (Formulate Initial Relationships, Patterns, and Structures in the Data), Forschungsfragen stellen (Pose Research Questions), Identifizieren relevante Daten für die Erklärung/ Argumentation (Identify Relevant Data for Explanation/Argumentation), Reasoning (Reasoning), Wechseln die Darstellung der Daten (Change the Representation of Data) \\
[4pt]
\bottomrule
\end{tabular}
\caption{%
  Scientific-inquiry Skill (\tasksk) item names per subject.
  Each entry is an inquiry competency that students are scored on independently.
}
\label{tab:sk-names}
\end{table*}

\FloatBarrier

\section{Rubric Entailment Experiments} \label{appendix:rubentail}
We experimented with both pretrained natural language inference (NLI) models and LLM prompting for this task.
After inspecting the results, we decided to use LLM prompting. The model used is \textit{GPT-4o-mini}.

\paragraph{Validation and motivation for LLM-based labelling.}
We validated the approach through manual inspection of a subset of rubric pairs.
Off-the-shelf NLI models available on Hugging Face (e.g., models fine-tuned on SNLI or MultiNLI) consistently struggled to capture the entailment structure specific to educational rubrics.
The core difficulty is that rubric levels do not express propositional claims about the world, but rather \emph{criteria for student performance}: a higher level such as \textit{``targeted use of a concept''} does not semantically entail a lower level such as \textit{``use without content''} in the linguistic sense that standard NLI models are trained to detect.
Instead, the relevant entailment is \emph{behavioural}: a student who satisfies a higher criterion necessarily also satisfies the weaker conditions of a lower one.
Standard NLI models, trained on news and Wikipedia inference pairs, lack exposure to this criterion-based reasoning pattern and systematically misclassify pairs that human experts and \textit{GPT-4o-mini} agree are entailing.
We therefore adopted LLM prompting with an explicit task-specific definition of entailment (see prompt below), which produced results consistent with our manual inspection of the labelled pairs.

\begin{samepage}
\begin{lstlisting}[
  title={Rubric Entailment Prompt},
  basicstyle=\ttfamily\small,
  breaklines=true,
  frame=single,
  backgroundcolor=\color{gray!10},
  xleftmargin=2em,
  framexleftmargin=1.5em,
  columns=fullflexible,
  keepspaces=true
]
You are an expert in educational rubric design. Your task is to determine whether
a higher-level rubric criterion logically entails a lower-level rubric criterion.

DEFINITION OF ENTAILMENT
A higher-level rubric entails a lower-level rubric if and only if every student
who satisfies the higher-level criterion NECESSARILY also satisfies the
lower-level criterion.

Example (entails = true):
  Lower (partial):  "Student describes X or Y."
  Higher (full):    "Student describes X and Y."
  -> Any student who does X and Y automatically also does X or Y. -> entails.

Example (entails = false):
  Lower level 1:    "Student mentions the concept without context."
  Higher level 3:   "Student applies the concept correctly in context."
  -> A student who applies it correctly need not have *merely* mentioned it
    without context first; the categories are qualitatively different. -> does NOT entail.

IMPORTANT
- Respond ONLY with a valid JSON object, no markdown fences.
- Format: {"reasoning": "<1-2 sentences>", "entails": true/false}
\end{lstlisting}
\end{samepage}

\FloatBarrier

\section{Per-Subject Performance} \label{appendix:per-subject}

\begin{figure}[h]
    \centering
    \includegraphics[width=\linewidth]{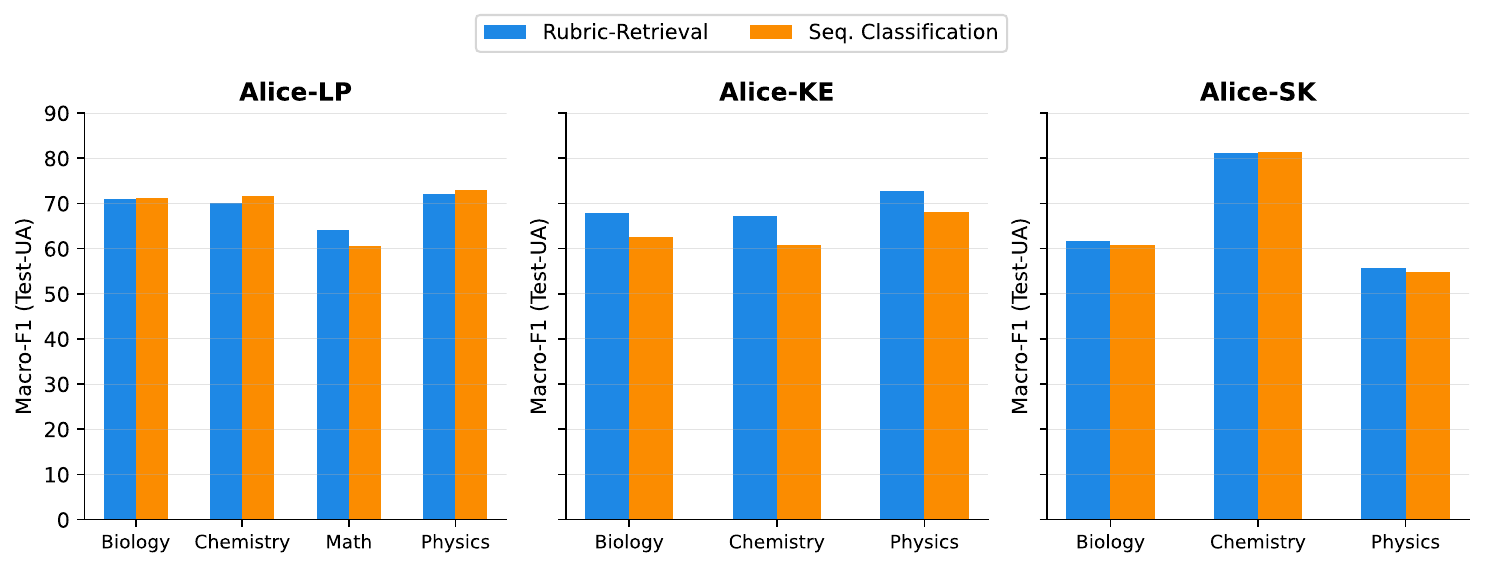}
    \caption{Per-subject Macro-F1 on Test-UA for each subtask, averaged across all input format variants within each approach. Fine-tuned models only; zero-shot LLM prompting was not evaluated on Test-UA.}
    \label{fig:per-subject-f1-ua}
\end{figure}

\begin{figure}[h]
    \centering
    \includegraphics[width=\linewidth]{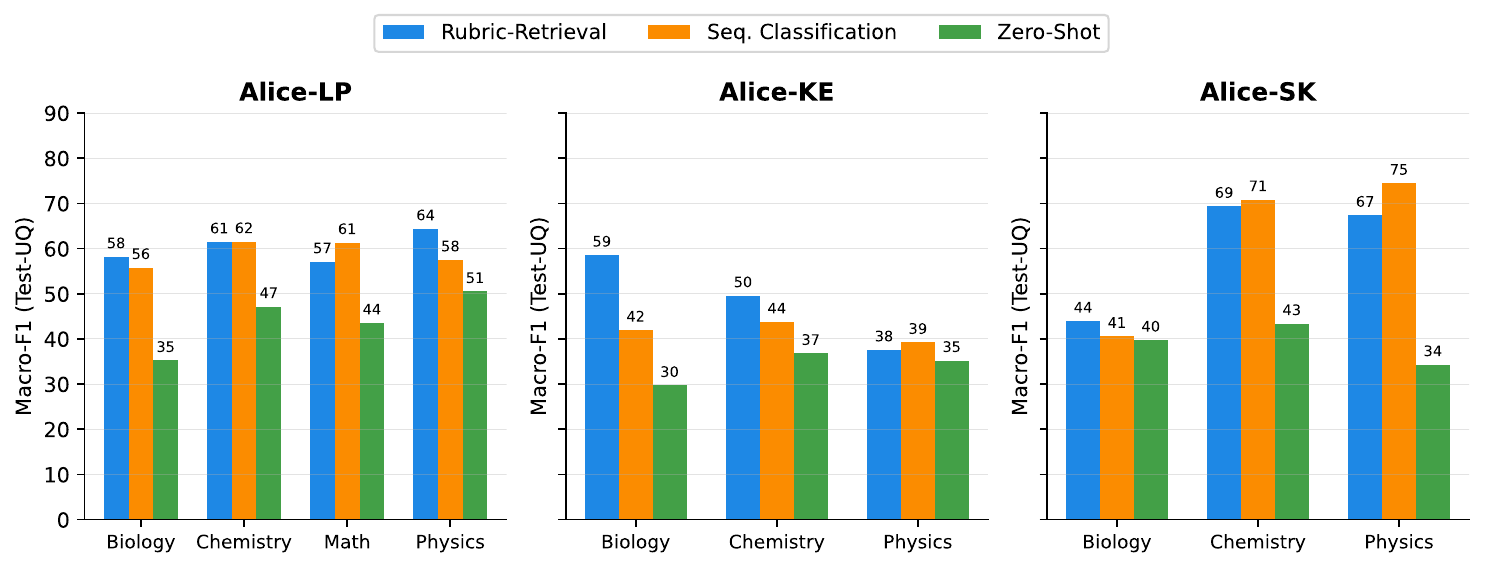}
    \caption{Per-subject Macro-F1 on Test-UQ for each subtask, averaged across all input format variants within each approach.}
    \label{fig:per-subject-f1}
\end{figure}

\FloatBarrier

\section{Model Configuration and Hyperparameters} \label{appendix:hyperparameters}

We randomly sample 10\% of the training data as a validation set and evaluate the model on it at each epoch. The model checkpoint that achieves the highest validation accuracy is saved for testing. Since the validation set is sampled instance-wise from the Train split (\autoref{tab:stats}), and the Train split is itself drawn only from the 90 questions not held out for Test-UQ (\autoref{sec3:dataset-statistics}), the validation set is question-disjoint from Test-UQ by construction; it is, however, instance-sampled with respect to Test-UA, as both are drawn from the same 90-question pool.

For the fine-tuned experiments we use XLM-R-Longformer-base-4096, mmBERT-base, and Llama-3.2-1B, 1B-Instruct, 3B, and 3B-Instruct. For zero-shot prompting we use Mistral-7B-Instruct-v0.3, Llama-3.1-8B-Instruct, Llama-3.3-70B-Instruct, GPT-4o-mini, and GPT-5-mini, all queried with greedy decoding (temperature $0.0$, top-$p$ $0.95$).

\textbf{MLM-based models} We fine-tune with a learning rate of $2\times10^{-5}$ for four epochs with a batch size of 16.

\textbf{LLM-based models} We train with 4-bit quantisation (QLoRA), a LoRA rank of 64, and a learning rate of $1\times10^{-4}$. We use a per-device batch size of 4 and gradient accumulation over 8 steps. All subtasks are trained for three epochs.

\FloatBarrier

\section{Additional QWK Results} \label{appendix:qwk-results}

\begin{table*}[t]
\centering
\small
\setlength{\tabcolsep}{3pt}

\begin{subtable}[t]{0.49\textwidth}
\centering
\caption{UA QWK}
\resizebox{\linewidth}{!}{%
\begin{tabular}{llcc|cc|cc}
\toprule
Approach & Model & \multicolumn{2}{c}{\alicelp} & \multicolumn{2}{c}{\aliceke} & \multicolumn{2}{c}{\alicesk} \\
\cmidrule(lr){3-4}\cmidrule(lr){5-6}\cmidrule(lr){7-8}
Rub. Ret. & Input format & \textit{ar} & \textit{+qs} & \textit{ar} & \textit{+qs} & \textit{ar} & \textit{+qs} \\
\midrule
 & XLM-RoBERTa-Long & 62.6 & \cellcolor{red!15}57.6 & 76.7 & \cellcolor{red!15}65.6 & 55.6 & 56.3 \\
 & mmBERT & 69.7 & \cellcolor{red!15}67.9 & 81.1 & 81.2 & 64.5 & 65.5 \\
 & Llama-3.2-1B & 69.5 & 72.9 & 81.4 & 82.2 & 64.7 & \cellcolor{red!15}64.2 \\
 & Llama-3.2-1B-Instruct & 70.5 & 73.2 & 81.3 & 82.9 & 64.6 & 67.9 \\
 & Llama-3.2-3B & 73.2 & 76.2 & 83.0 & 83.7 & 62.5 & 66.3 \\
 & Llama-3.2-3B-Instruct & 74.4 & 77.7 & 83.3 & 83.3 & 64.4 & 69.1 \\
\midrule
Seq. Class. & Input format & \textit{sa} & \textit{+qr} & \textit{sa} & \textit{+qr} & \textit{sa} & \textit{+qr} \\
\midrule
 & XLM-RoBERTa-Long & \underline{65.4} & \cellcolor{red!15}\underline{59.9} & 75.4 & \underline{76.6} & \underline{62.2} & \cellcolor{red!15}\underline{61.9} \\
 & mmBERT & \underline{73.3} & \cellcolor{red!15}\underline{72.1} & 54.1 & \underline{81.8} & 62.0 & 65.3 \\
 & Llama-3.2-1B & \underline{73.1} & \underline{73.6} & \underline{81.6} & \cellcolor{red!15}81.4 & 61.3 & 63.7 \\
 & Llama-3.2-1B-Instruct & \underline{75.7} & \cellcolor{red!15}\underline{75.3} & 54.4 & 81.4 & 62.7 & 67.7 \\
 & Llama-3.2-3B & \underline{77.0} & \cellcolor{red!15}\underline{76.6} & 55.6 & \underline{83.7} & \underline{63.0} & \underline{68.8} \\
 & Llama-3.2-3B-Instruct & \underline{\textbf{78.0}} & \cellcolor{red!15}77.0 & 55.6 & \underline{\textbf{83.8}} & 61.7 & \underline{\textbf{69.8}} \\
\bottomrule
\end{tabular}
}
\end{subtable}
\hfill
\begin{subtable}[t]{0.49\textwidth}
\centering
\caption{UQ QWK}
\resizebox{\linewidth}{!}{%
\begin{tabular}{llcc|cc|cc}
\toprule
Approach & Model & \multicolumn{2}{c}{\alicelp} & \multicolumn{2}{c}{\aliceke} & \multicolumn{2}{c}{\alicesk} \\
\cmidrule(lr){3-4}\cmidrule(lr){5-6}\cmidrule(lr){7-8}
Rub. Ret. & Input format & \textit{ar} & \textit{+qs} & \textit{ar} & \textit{+qs} & \textit{ar} & \textit{+qs} \\
\midrule
 & XLM-RoBERTa-Long & 57.1 & \cellcolor{red!15}47.8 & 47.2 & \cellcolor{red!15}25.7 & 38.7 & \cellcolor{red!15}32.5 \\
 & mmBERT & 50.8 & 57.5 & 51.0 & 54.9 & 37.0 & 45.3 \\
 & Llama-3.2-1B & 57.5 & 65.4 & 60.5 & \cellcolor{red!15}60.1 & 40.5 & 45.9 \\
 & Llama-3.2-1B-Instruct & 61.8 & 66.0 & 59.8 & \cellcolor{red!15}57.6 & 38.4 & 43.7 \\
 & Llama-3.2-3B & 59.5 & 62.6 & 54.5 & 64.5 & 43.3 & 52.7 \\
 & Llama-3.2-3B-Instruct & 53.3 & \textbf{69.6} & 55.9 & 64.6 & 44.1 & \textbf{56.0} \\
\midrule
Seq. Class. & Input format & \textit{sa} & \textit{+qr} & \textit{sa} & \textit{+qr} & \textit{sa} & \textit{+qr} \\
\midrule
 & XLM-RoBERTa-Long & 53.1 & \cellcolor{red!15}\underline{52.1} & 47.2 & \cellcolor{red!15}\underline{45.9} & 37.9 & \underline{42.6} \\
 & mmBERT & \underline{60.2} & \cellcolor{red!15}56.7 & 39.7 & \underline{61.9} & 34.2 & 39.8 \\
 & Llama-3.2-1B & \underline{60.4} & \cellcolor{red!15}57.7 & 56.1 & 58.7 & 34.5 & 40.9 \\
 & Llama-3.2-1B-Instruct & \underline{63.1} & 64.5 & 39.5 & \underline{59.1} & \underline{38.5} & \underline{48.0} \\
 & Llama-3.2-3B & \underline{65.5} & \underline{66.3} & 39.0 & 57.9 & 41.7 & \underline{53.8} \\
 & Llama-3.2-3B-Instruct & \underline{66.2} & \cellcolor{red!15}64.0 & 40.8 & \underline{\textbf{64.8}} & \underline{47.6} & 50.5 \\
\bottomrule
\end{tabular}
}
\end{subtable}

\vspace{0.5em}

\begin{subtable}[t]{0.74\textwidth}
\centering
\caption{Zero-shot UQ QWK}
\scriptsize
\setlength{\tabcolsep}{3pt}
\begin{tabular}{lccc|ccc|ccc}
\toprule
Model & \multicolumn{3}{c}{\tasklp} & \multicolumn{3}{c}{\taskke} & \multicolumn{3}{c}{\tasksk} \\
\cmidrule(lr){2-4}\cmidrule(lr){5-7}\cmidrule(lr){8-10}
 & \textit{-r} & \textit{+r} & \textit{+qs+r} & \textit{-r} & \textit{+r} & \textit{+qs+r} & \textit{-r} & \textit{+r} & \textit{+qs+r} \\
\midrule
Mistral-7B-Instruct & 16.4 & 42.3 & 33.1 & 42.1 & 40.1 & \cellcolor{red!15}39.1 & 20.5 & 22.6 & 27.4 \\
Llama-3.1-8B-Instruct & \textbf{34.7} & 35.8 & \cellcolor{red!15}32.7 & 34.3 & 36.5 & \cellcolor{red!15}26.7 & 20.8 & 16.8 & \cellcolor{red!15}3.4 \\
Llama-3.3-70B-Instruct & 31.2 & 56.1 & 55.7 & 49.8 & 54.2 & \cellcolor{red!15}44.8 & \textbf{27.1} & \textbf{33.5} & \textbf{37.3} \\
GPT-4o-mini & 25.0 & 55.2 & \textbf{59.6} & 46.4 & 55.6 & 50.4 & 20.6 & 19.6 & 25.2 \\
GPT-5-mini & 32.1 & \textbf{67.0} & -- & \textbf{62.0} & \textbf{67.8} & \textbf{70.4} & 21.8 & 23.8 & 33.4 \\
\bottomrule
\end{tabular}
\end{subtable}

\caption{Complementary QWK results on \textsc{Alice}. The table follows the same layout as \autoref{tab:main-results}: upper panels show fine-tuned UA and UQ results, while the lower panel shows zero-shot UQ prompting. In the upper panels, bold marks the best result per subtask; in the lower panel, it marks the best result per input format and subtask. The \underline{underline} in the upper panels marks cells where sequence classification outperforms rubric-retrieval. \colorbox{red!15}{Red cells} mark full-context results (\textit{+qs}/\textit{+qr}/\textit{+qs+r}) that underperform the base input format (\textit{ar}/\textit{sa}/\textit{-r}) for that model and subtask. Dashes indicate unavailable QWK values.}
\label{tab:qwk-results}
\end{table*}

\FloatBarrier

\section{Ablation Studies} \label{appendix:lp-ablation}
We run two additional ablations on \alicelp: (i) training with a contrastive loss (CL), and (ii) a bi-encoder variant that encodes answers and rubric items separately.

\paragraph{Contrastive loss}
For each student--rubric pair $(a_i, r_j^i)$ with binary label $y_{i,j} \in \{0,1\}$, we use the contrastive objective:
\begin{equation}
\ell_{\text{contrast}}^{(i,j)} =
\begin{cases}
(1 - z_{i,j})^2 & \text{if } y_{i,j} = 1 \\
(z_{i,j})^2     & \text{if } y_{i,j} = 0
\end{cases}
\end{equation}

\autoref{tab:lp-cl-only} reports CL results under the same model and input settings (\textit{ar} and \textit{+qs}) as the main experiments.

\begin{table}[ht]
\centering
\small
\resizebox{\linewidth}{!}{
\begin{tabular}{llcc}
\toprule
Base Model & Input Format & UA (F1/Acc) & UQ (F1/Acc) \\
\midrule
\multirow{2}{*}{\textit{XLM-RoBERTa-Long}} & ar  &  65.7 / 65.5 & 61.0 / 60.7 \\
 & +qs & \cellcolor{gray!20} 65.0 / 64.8 & \cellcolor{gray!20} 59.4 / 59.1 \\
\hline
\multirow{2}{*}{\textit{mmBERT}} & ar  & 69.7 / 69.5 & 57.7 / 57.5 \\
 & +qs & 70.0 / 69.8 & 60.0 / 60.0 \\
\hline
\multirow{2}{*}{\textit{Llama-3.2-1B}} & ar  & 72.1 / 71.9 & 59.0 / 58.8 \\
 & +qs & 73.5 / 73.2 & 59.9 / 59.5 \\
\hline
\multirow{2}{*}{\textit{Llama-3.2-1B-Instruct}} & ar  & 71.5 / 71.2 & 61.7 / 61.3 \\
 & +qs & 72.8 / 72.4 & 60.6 / 60.4 \\
\hline
\multirow{2}{*}{\textit{Llama-3.2-3B}} & ar  & \cellcolor{gray!20} 73.2 / 72.8 & 59.5 / 59.1 \\
 & +qs & 75.8 / 75.5 & 63.2 / 62.5 \\
\hline
\multirow{2}{*}{\textit{Llama-3.2-3B-Instruct}} & ar  & \cellcolor{gray!20} 73.7 / 73.4 & 61.0 / 60.7 \\
 & +qs & 75.4 / 75.0 & 65.9 / 65.1 \\
\bottomrule
\end{tabular}
}
\caption{Contrastive loss (CL) results on \alicelp. \colorbox{gray!20}{Gray cells} indicate CL results that outperform the corresponding rubric-retrieval setting in \autoref{tab:main-results} based on F1.}
\label{tab:lp-cl-only}
\end{table}

\paragraph{Separate Encoding}
We encode the student answer and each rubric element independently using the same LM encoder. For the +qs input format, the answer is encoded jointly with the question and/or sample solution, while rubric elements are always encoded in isolation. Answer and rubric embeddings are combined via element-wise multiplication.
We use SCE as the loss function.

\autoref{tab:lp-separate-only} reports separate-encoding results under the same model and input settings (\textit{ar} and \textit{+qs}) as the main experiments.

\begin{table}[ht]
\centering
\small
\resizebox{\linewidth}{!}{
\begin{tabular}{llcc}
\toprule
Base Model & Input Format & UA (F1/Acc) & UQ (F1/Acc) \\
\midrule
\multirow{2}{*}{\textit{XLM-RoBERTa-Long}} & ar   & 66.4 / 66.0 & 56.6 / 56.3 \\
 & +qs & 63.6 / 63.5 & \cellcolor{gray!20} 57.4 / 57.3 \\
\hline
\multirow{2}{*}{\textit{mmBERT}} & ar   & 68.6 / 68.1 & 57.1 / 57.0 \\
 & +qs & \cellcolor{gray!20} 70.4 / 70.0 & \cellcolor{gray!20} 60.2 / 59.7 \\
\hline
\multirow{2}{*}{\textit{Llama-3.2-1B}} & ar   & 70.0 / 69.6 & 57.8 / 57.9 \\
 & +qs & \cellcolor{gray!20} 73.6 / 73.2 & 61.6 / 61.0 \\
\hline
\multirow{2}{*}{\textit{Llama-3.2-1B-Instruct}} & ar   & 70.0 / 69.6 & 56.9 / 56.8 \\
 & +qs & \cellcolor{gray!20} 74.0 / 73.6 & \cellcolor{gray!20} 66.2 / 65.6 \\
\hline
\multirow{2}{*}{\textit{Llama-3.2-3B}} & ar   & 72.9 / 72.5 & 57.9 / 58.1 \\
 & +qs & \cellcolor{gray!20} 75.8 / 75.4 & 64.2 / 63.4 \\
\hline
\multirow{2}{*}{\textit{Llama-3.2-3B-Instruct}} & ar   & 70.9 / 70.5 & 56.7 / 56.8 \\
 & +qs & 75.5 / 75.2 & 62.4 / 62.1 \\
\bottomrule
\end{tabular}
}
\caption{Separate encoding results on \alicelp. \colorbox{gray!20}{Gray cells} indicate separate-encoding results that outperform the corresponding rubric-retrieval setting in \autoref{tab:main-results} based on F1.}
\label{tab:lp-separate-only}
\end{table}

Overall, both CL and the bi-encoder variant are competitive on Test-UA (unseen answers), with several settings matching or exceeding the main \tasklp configuration. However, on Test-UQ (unseen questions), they generally underperform compared with our proposed rubric-retrieval setup, despite a few model-specific gains. This pattern suggests that our proposed modelling approach generalises better to new questions.

\FloatBarrier

\section{Zero-shot LLM Prompts} \label{appendix:llms}

\lstset{
  basicstyle=\ttfamily\small,
  breaklines=true,
  frame=single,
  backgroundcolor=\color{gray!10},
  xleftmargin=2em,
  framexleftmargin=1.5em,
  columns=fullflexible,
  keepspaces=true
}

\FloatBarrier
\subsection{System Prompts}
For each subtask, the system prompt combines task-specific instructions, common scoring rules, and a task-appropriate output schema. The German prompt components used in the experiments are shown together with English translations.

\begin{lstlisting}[caption={Task-Specific Instructions (German)}]
LP
Sie werden die Antwort eines K12-Schuelers in MINT-Faechern bewerten. ZIEL
Weisen Sie basierend auf dem bereitgestellten Rubrikensatz genau EINE Rubrikstufen-Bezeichnung zu. Sie muessen die Schuelerantwort intern anhand jeder Rubrik bewerten und dann die am besten passende Bezeichnung auswaehlen.

FOKUS: Bewerten Sie die GESAMTLEISTUNG des Schuelers bei der Bearbeitung der Aufgabe, einschliesslich Verstaendnis, Anwendung und Kommunikation des Wissens.

KE
Sie werden die Antwort eines K12-Schuelers in MINT-Faechern bewerten. ZIEL
Bewerten Sie JEDES bereitgestellte Wissenselement und vergeben Sie fuer jedes Element genau EINE Rubrikstufen-Bezeichnung.

FOKUS: Bewerten Sie die BEHERRSCHUNG SPEZIFISCHER WISSENSELEMENTE (z.B. chemische Reaktionen, physikalische Gesetze, mathematische Konzepte). Konzentrieren Sie sich auf die Korrektheit und Tiefe des fachlichen Verstaendnisses der jeweiligen Wissenskomponente.

SK
Sie werden die Antwort eines K12-Schuelers in MINT-Faechern bewerten. ZIEL
Bewerten Sie JEDE bereitgestellte Faehigkeit und vergeben Sie fuer jede Faehigkeit genau EINE Rubrikstufen-Bezeichnung.

FOKUS: Bestimmen Sie, ob die jeweilige Faehigkeit in der Antwort des Schuelers VORHANDEN oder NICHT VORHANDEN ist (z.B. Hypothesen formulieren, Experimente entwerfen, Daten interpretieren, Schlussfolgerungen ziehen). Verwenden Sie die passende Bezeichnung aus dem bereitgestellten Rubrikensatz.
\end{lstlisting}

\begin{lstlisting}[caption={Task-Specific Instructions (English Translation)}]
LP
You will evaluate the answer of a K-12 student. GOAL
Assign exactly ONE rubric level label based on the provided rubric. You must internally evaluate the student answer against each rubric level, then select the best matching label.

FOCUS: Evaluate the student's overall performance on the task, including understanding, application, and communication of knowledge.

KE
You will evaluate the answer of a K-12 student. GOAL
Evaluate EACH provided knowledge element and assign exactly ONE rubric level label for each element.

FOCUS: Evaluate mastery of specific knowledge elements, such as scientific concepts, mathematical ideas, or domain facts. Focus on the correctness and depth of understanding for each knowledge component.

SK
You will evaluate the answer of a K-12 student. GOAL
Evaluate EACH provided skill and assign exactly ONE rubric level label for each skill.

FOCUS: Determine whether each skill is PRESENT or ABSENT in the student's answer (e.g. formulating hypotheses, designing experiments, interpreting data, drawing conclusions). Use the matching label from the provided rubric.
\end{lstlisting}

\begin{lstlisting}[caption={Common Scoring Rules (German)}]
REGELN
- Beruecksichtigen Sie die Frage und die Beispielantworten nur, wenn sie bereitgestellt werden; sie sind optionale Hilfsmittel zum Verstaendnis, keine strikten Anforderungen.
- Eine Rubrik ist nur erfuellt, wenn ihre erforderlichen Kriterien erfuellt sind. Wenn eine Rubrik "eines von/von mehreren" verwendet, folgen Sie dieser Logik.
- Bewerten Sie ausschliesslich basierend auf der Uebereinstimmung zwischen der Schuelerantwort und den Rubrikkriterien (und der Frage, falls vorhanden). Ignorieren Sie den Stil, es sei denn, er wird von der Rubrik verlangt.
- Geben Sie als Bewertung ausschliesslich die exakte Bezeichnung der am besten passenden Rubrikstufe zurueck, nicht deren numerische ID.
\end{lstlisting}

\begin{lstlisting}[caption={Common Scoring Rules (English Translation)}]
RULES
- Consider the question and sample answers only when they are provided; they are optional aids for understanding, not strict requirements.
- A rubric level is satisfied only when its required criteria are met. If a rubric uses "one of" or "several of" logic, follow that logic.
- Evaluate only based on the match between the student answer and the rubric criteria, and the question if provided. Ignore style unless the rubric requires it.
- Return the exact label name of the best matching rubric level as the evaluation, not its numeric ID.
\end{lstlisting}

\begin{lstlisting}[caption={Output Schemas (German)}]
LP
AUSGABE
Geben Sie NUR ein gueltiges JSON-Objekt mit genau diesen Feldern zurueck:
"reasoning": "1-3 praegnante Saetze, die erklaeren, warum diese Bezeichnung am besten zur Rubrik passt."
"score": "<exakte Rubrikstufen-Bezeichnung>"

WICHTIG
- Verwenden Sie fuer "score" ausschliesslich eine Rubrikstufen-Bezeichnung aus dem bereitgestellten Rubrikensatz, keine Zahl.
- Fuegen Sie KEINE zusaetzlichen Schluessel, Texte oder Formatierungen ausserhalb des JSON hinzu.

KE/SK
AUSGABE
Geben Sie NUR ein gueltiges JSON-Objekt mit genau diesen Feldern zurueck:
"reasoning": "1-3 praegnante Saetze fuer die Gesamtbegruendung."
"scores": {"<Elementname>": "<exakte Rubrikstufen-Bezeichnung>", "<AnotherElement>": "<exakte Rubrikstufen-Bezeichnung>", ...}

WICHTIG
- Verwenden Sie "scores" mit Elementnamen als Schluessel, unabhaengig von der Anzahl der Elemente.
- Das Objekt "scores" muss jedes bereitgestellte Element genau einmal enthalten.
- Verwenden Sie als Werte ausschliesslich Rubrikstufen-Bezeichnungen aus dem bereitgestellten Rubrikensatz, keine Zahlen.
- Fuegen Sie KEINE zusaetzlichen Schluessel ausser "reasoning" und "scores" hinzu.
\end{lstlisting}

\begin{lstlisting}[caption={Output Schemas (English Translation)}]
LP
OUTPUT
Return ONLY a valid JSON object with exactly these fields:
"reasoning": "1-3 concise sentences explaining why this label best matches the rubric."
"score": "<exact rubric level label>"

IMPORTANT
- Use only a rubric level label from the provided rubric for "score", not a number.
- Do NOT add extra keys, text, or formatting outside the JSON object.

KE/SK
OUTPUT
Return ONLY a valid JSON object with exactly these fields:
"reasoning": "1-3 concise sentences for the overall rationale."
"scores": {"<ElementName>": "<exact rubric level label>", "<AnotherElement>": "<exact rubric level label>", ...}

IMPORTANT
- Use "scores" with element names as keys, regardless of how many elements there are.
- The "scores" object must contain every provided element exactly once.
- Use only rubric level labels from the provided rubric as values, not numbers.
- Do NOT add keys other than "reasoning" and "scores".
\end{lstlisting}

\subsection{User Prompt}

\begin{lstlisting}[caption={User Prompt (German)}]
{element_section}--- RUBRIKENSATZ ---
{rubrics}
{question_section}{sample_solutions_section}--- SCHUELERANTWORT ---
{answer}
\end{lstlisting}

\begin{lstlisting}[caption={User Prompt (English Translation)}]
{element_section}--- RUBRIC ---
{rubrics}
{question_section}{sample_solutions_section}--- STUDENT ANSWER ---
{answer}
\end{lstlisting}

\FloatBarrier
\section{Benchmarking on ASAP-SAS} \label{appendix:asap-sas}
We also benchmarked our method on \textsc{ASAP-SAS}, which covers diverse domains such as English literature and science~\citep{bexte-etal-2022-similarity}. For each model, we experiment with both the basic \textit{ar} and \textit{+q} input formats. All listed previous methods train a \textbf{separate model} for each question. Below are short descriptions of each method.
\begin{table*}[t]
\centering
\resizebox{1.0\textwidth}{!}{%
\begin{tabular}{l|l|cccccccccc|c}
\toprule
\multicolumn{13}{c}{\textbf{Joint Rubric-Retrieval}} \\
\midrule
Base Model & Input Format & Q1 & Q2 & Q3 & Q4 & Q5 & Q6 & Q7 & Q8 & Q9 & Q10 & Mean \\
\midrule
\multirow{2}{*}{\textit{BERT-base-cased}} & ar & 81.5 & 77.4 & 67.6 & 70.9 & 78.6 & 81.9 & 67.6 & 62.9 & 78.1 & 71.4 & 73.8 \\
& +q & 77.2 & 62.0 & 66.1 & 69.5 & 63.3 & 58.7 & 66.3 & 63.6 & 82.8 & 73.8 & 68.3 \\
\cmidrule{1-13}
\multirow{2}{*}{\textit{RoBERTa-base}} & ar & 83.3 & 72.0 & 68.9 & 76.0 & 77.2 & 83.2 & 70.3 & 69.4 & 79.8 & 75.5 & 75.5 \\
& +q & 81.1 & 67.5 & 66.4 & 71.9 & 68.0 & 66.5 & 69.0 & 67.1 & 81.3 & 76.9 & 71.6 \\
\cmidrule{1-13}
\multirow{2}{*}{\textit{ModernBERT-base}} & ar & 76.3 & 57.5 & 68.3 & 65.7 & 73.4 & 75.6 & 56.8 & 52.5 & 79.6 & 70.2 & 67.6 \\
& +q & 72.9 & 58.3 & 67.7 & 65.6 & 76.4 & 74.5 & 58.5 & 56.8 & 75.9 & 69.6 & 67.6 \\
\cmidrule{1-13}
\multirow{2}{*}{\textit{XLM-RoBERTa-Long}} & ar & 80.9 & 70.8 & 69.8 & 73.1 & 82.7 & 84.2 & 67.7 & 56.8 & 79.2 & 74.4 & 73.9 \\
& +q & 74.0 & 59.2 & 67.2 & 64.1 & 62.1 & 72.5 & 62.0 & 63.3 & 82.4 & 74.5 & 68.2 \\
\midrule
\multirow{2}{*}{\textit{Llama-3.2-1B}} & ar & 84.6 & 80.9 & 69.5 & \textbf{77.0} & 81.0 & 80.7 & 72.0 & 67.7 & 82.8 & 76.3 & 77.2 \\
& +q & 85.6 & 80.6 & 65.8 & 73.8 & 80.5 & 83.9 & 71.7 & 71.5 & 82.4 & 77.5 & 77.3 \\
\cmidrule{1-13}
\multirow{2}{*}{\textit{Llama-3.2-1B-Instruct}} & ar & 81.5 & 77.8 & 68.5 & 74.7 & 82.9 & 83.5 & 69.1 & 68.9 & 82.0 & 76.4 & 76.5 \\
& +q & 85.5 & 80.1 & 67.1 & 70.9 & 82.1 & 85.8 & 67.5 & 71.6 & 81.3 & 75.7 & 76.8 \\
\cmidrule{1-13}
\multirow{2}{*}{\textit{Llama-3.2-3B}} & ar & 85.8 & 80.7 & 71.3 & 69.5 & 83.7 & 86.3 & 71.6 & 71.3 & 83.9 & 77.1 & 78.1 \\
& +q & 87.8 & \textbf{86.3} & 69.3 & 74.2 & \textbf{85.2} & 85.0 & 70.3 & 70.9 & \textbf{84.3} & 77.1 & 79.0 \\
\cmidrule{1-13}
\multirow{2}{*}{\textit{Llama-3.2-3B-Instruct}} & ar & 87.8 & 83.5 & 67.4 & 69.2 & 82.9 & 85.4 & 71.4 & 72.1 & 83.1 & 79.2 & 78.2 \\
& +q & 87.7 & 84.4 & 71.2 & 73.9 & 84.5 & 85.2 & \textbf{74.8} & 70.6 & 81.3 & 77.3 & \textbf{79.1} \\
\midrule
\multicolumn{13}{c}{\textbf{Previous Methods}} \\
\midrule
Method & & Q1 & Q2 & Q3 & Q4 & Q5 & Q6 & Q7 & Q8 & Q9 & Q10 & Mean \\
\midrule
\citet{ramachandran-etal-2015-identifying} & & 86.0 & 78.0 & 66.0 & 70.0 & 84.0 & \textbf{88.0} & 66.0 & 63.0 & 84.0 & 79.0 & 78.0 \\
\citet{riordan-etal-2017-investigating} & & 79.5 & 71.8 & 68.4 & 70.0 & 83.0 & 79.0 & 64.8 & 55.4 & 77.7 & 73.5 & 72.3 \\
\citet{wang-etal-2019-inject-rubric} & & 79.2 & 71.4 & NA & NA & 80.4 & 79.3 & NA & NA & NA & NA &  NA \\
\citet{kumar2019get} & & 87.2 & 82.4 & \textbf{74.5} & 74.3 & 84.5 & 85.8 & 72.5 & 62.4 & \textbf{84.3} & \textbf{83.2} &  \textbf{79.1} \\
\citet{bexte-etal-2022-similarity} MAX & & 84.0 & 79.0 & 67.0 & 67.0 & 72.0 & 58.0 & 68.0 & \textbf{77.0} & 72.0 & 72.0 & 72.4 \\
\citet{bexte-etal-2022-similarity} AVG & & \textbf{88.0} & 78.0 & \textbf{74.0} & 72.0 & 69.0 & 74.0 & 58.0 & 69.0 & 77.0 & 72.0 & 73.1 \\
\bottomrule
\end{tabular}%
}

\caption{Comparison of our rubric-retrieval framework with previous results on \textsc{ASAP-SAS} (QWK, \%).}
\label{tab:asap-sas}
\end{table*}
\paragraph{\citet{ramachandran-etal-2015-identifying}}
The paper proposes a supervised short-answer scoring approach that represents student responses using a combination of lexical overlap, semantic similarity (via WordNet and distributional semantics), and question-specific reference answers, which are then fed into a regression/classification model. The method is deliberately feature-driven rather than neural, reflecting the low-resource and interpretability constraints of the time.
\paragraph{\textbf{\citet{kumar2019get}}}
The paper proposes AutoSAS, a supervised short-answer scoring system that predicts a numeric score for a student response using engineered linguistic and semantic features. For each answer, the system extracts lexical diversity measures, Word2Vec embeddings, and overlap features between the student response and the prompt or reference content. These features are then fed into a regression model that outputs a score. They built a separate model for each question.
\paragraph{\textbf{\citet{wang-etal-2019-inject-rubric}}}
It proposes a neural short answer grading model that augments a baseline encoder with an explicit rubric component. The architecture has two parts: (1) a base component that encodes student answers using a BiLSTM over word embeddings to produce a feature vector, and (2) a rubric component that computes word-level attention alignments between the answer and each key element in the rubric to produce rubric-aware features. These features are merged (by concatenation or weighted sum) and passed through a regression layer to predict the score, allowing the model to learn alignment between rubric criteria and answer content.
\paragraph{\textbf{\citet{riordan-etal-2017-investigating}}}
The paper evaluates simple neural architectures for short-answer scoring by training CNN- and LSTM-based models on student responses, using word-embedding inputs and treating scoring as regression or classification depending on the dataset. The models take only the student's answer text (plus prompt-specific training data) and do not incorporate rubrics or reference criteria as structured inputs. They compare these neural models to a strong feature-engineered baseline across three SAS datasets and find that neural models can outperform the baseline, though the best architecture varies with prompt characteristics. They built a separate model for each question.

\paragraph{\textbf{\citet{bexte-etal-2022-similarity}}}
The authors train a model that predicts whether a pair of answers belongs to the same level. During testing, the model uses the maximum or average similarity between a candidate answer and anchor answers from different levels to determine the final score.

\FloatBarrier
\autoref{tab:asap-sas} reports per-question QWK scores on ASAP-SAS following the official evaluation protocol. Overall, our joint rubric-retrieval models achieve competitive performance compared to prior work that trains separate models for each prompt. The LLM encoders outperform the MLM encoders on average and, unlike the MLMs, generally maintain or improve their mean performance when the question prompt is added. This is consistent with the finding in \autoref{sec5:results} that learning-performance scoring benefits from LLM encoders.

Although some previous methods use smaller models (e.g.\ \textit{BERT}) and achieve performance comparable to ours, they rely on prompt-specific training and tuning, which deviates from realistic deployment scenarios where new questions are encountered without retraining.

In contrast, our approach uses a single jointly trained model across all prompts. Among our models, lightweight LLM encoders (e.g., \textit{Llama-3.2-3B} variants) achieve a mean QWK comparable to or exceeding most prior neural approaches, despite the more challenging joint-training setting. These results indicate that rubric-retrieval with joint training trades off maximal per-prompt performance for improved generality and practical applicability, aligning better with real-world ASAS use cases.

\end{document}